\documentclass[journal]{IEEEtran}

\usepackage[sorting=none,style=ieee]{biblatex}
\usepackage{enumerate}
\usepackage{caption}
\usepackage{subfig}
\usepackage{dirtytalk}
\usepackage{array}
\usepackage{xspace}
\usepackage{tipa}
\usepackage{mathtools}
\usepackage{graphicx}
\usepackage{graphbox} 
\usepackage{amssymb}
\usepackage{amsmath}
\usepackage{algorithm}
\usepackage[noend]{algpseudocode}
\usepackage[acronym]{glossaries}
\usepackage[utf8]{inputenc}
\usepackage[T1]{fontenc}
\usepackage{nomencl}
\usepackage[cal=dutchcal]{mathalfa}
\usepackage{pifont}
\newcommand{\cmark}{\ding{51}}
\newcommand{\xmark}{\ding{55}}

\makeglossaries

\newacronym{cnn}{CNN}{Convolutional Neural Networks}
\newacronym{gm}{GM}{Gradient Monitoring}
\newacronym{rl}{RL}{Reinforcement Learning}
\newacronym{ppo}{PPO}{Proximal Policy Optimization}
\newacronym{dnn}{DNN}{Deep Neural Networks}
\newacronym{gmrl}{GMRL}{Gradient Monitored Reinforcement Learning}
\newacronym{a2c}{A2C}{Advantage Actor Critic}

\makenomenclature

\graphicspath{ {./images/} }

\begin{document}

%
\title{ Gradient Monitored Reinforcement Learning\\}






\author{
\IEEEauthorblockN{Mohammed Sharafath Abdul Hameed \IEEEauthorrefmark{1}, Gavneet Singh Chadha \IEEEauthorrefmark{1}, Andreas Schwung \IEEEauthorrefmark{1}, Steven X. Ding \IEEEauthorrefmark{2}.}

\IEEEauthorblockA{\IEEEauthorrefmark{1} Department of Automation Technology\\
South Westphalia University of Applied Sciences\\
  Soest, Germany.\\
  \emph{sharafath.mohammed, chadha.gavneetsingh, schwung.andreas@fh-swf.de}}\\
\IEEEauthorblockA{\IEEEauthorrefmark{2}Department of Automatic Control and Complex Systems\\
  University of Duisburg-Essen  \\
  Duisburg, Germany.\\ \emph{steven.ding@uni-due.de}}
}

\date{today}
\maketitle

\begin{abstract}
This paper presents a novel neural network training approach for faster convergence  and better generalization abilities in deep reinforcement learning. Particularly, we focus on the enhancement of training and evaluation performance in reinforcement learning algorithms by systematically reducing  gradient's variance and thereby providing a more targeted learning process. The  proposed method which we term as Gradient Monitoring(GM), is an approach to steer the learning in the weight parameters of a  neural network based on the dynamic development and feedback from the  training process itself. We propose different variants of the GM methodology which have been proven to increase the underlying performance of the model. The one of the proposed variant, Momentum with Gradient Monitoring (M-WGM), allows for a continuous adjustment of the quantum of back-propagated gradients in the network based on certain learning parameters. We further enhance the method with  Adaptive Momentum with Gradient Monitoring (AM-WGM) method which allows for automatic adjustment between focused learning of certain weights versus a more dispersed  learning depending on the feedback from the rewards collected. As a by-product, it also allows for automatic derivation of the required deep network sizes during training as the algorithm automatically freezes trained weights. The approach is applied to two discrete (Multi-Robot Co-ordination problem and Atari games) and one continuous control task (MuJoCo) using Advantage Actor-Critic (A2C) and Proximal Policy Optimization (PPO) respectively. The results obtained particularly underline the applicability and performance improvements of the methods in terms of  generalization capability.
\end{abstract}


\mbox{}





\section{Introduction}

Research in deep \acrfull{rl} has seen tremendous progress in recent years with widespread success in various areas including video  games~\cite{Lample.2017}, board games~\cite{Silver.2017}, robotics~\cite{Gu.2017}, industrial assembly~\cite{T.Inoue.2017} and continuous control tasks~\cite{TimothyP.Lillicrap.2016} among others. This rapid increase in interest in the research community can be particularly traced back to advances made in the training of  \acrfull{dnn}  in the last decade, as well as novel \acrshort{rl} algorithms developed recently. Notable example of the latter include value function based methods like deep Q-networks~\cite{Mnih.2015}, policy gradient methods like deep deterministic policy gradient~\cite{TimothyP.Lillicrap.2016}, \acrfull{a2c}~\cite{Mnih.2016}, trust region policy optimization~\cite{Schulman.2015} and \acrfull{ppo}~\cite{Schulman.2017} to name a few. Also additional training components have helped in improving \acrshort{rl} capabilities like improved exploration strategies~\cite{Conti.2018}, intrinsic motivation~\cite{Mohamed.2015} and curiosity-driven methods~\cite{Pathak.2017}.

Revisiting the training of \acrshort{dnn}, regularization and better optimization methods have played a crucial role in improving their generalization capabilities,  where Batch Normalization~\cite{Ioffe.2015b}, Dropout~\cite{Srivastava.2014} and weight decay~\cite{Goodfellow.2016} are the most prominent examples which have become a standard in supervised learning. Surprisingly, little attention has been paid to methods for improving the generalization capabilities of \acrshort{dnn} during reinforcement learning, although this appears to be crucial in supervised and unsupervised learning tasks. Regardless, most of the above mentioned approaches are also utilized in \acrshort{rl}, although there are stark differences between supervised learning and \acrshort{rl}. It must be noted however that the above methods nevertheless also assist in \acrshort{rl} training~\cite{Cobbe.2019}. Our goal however, is to develop a principled optimization and training  approach for \acrshort{rl}, especially  considering its dynamic learning process. 

In literature, generalization in \acrshort{rl} is usually done by testing the trained agent's performance on an unseen variation of the environment, usually performed  by procedurally generating new  environments \cite{Cobbe.2019}.  We however, want to improve the evaluation performance on the same environment rather than generating new and unseen environments for the agent. An introduction to the existing methods for generalization in \acrshort{rl} is provided in Section~\ref{sec:relwork}. As a related problem, the derivation of suitable network sizes for a particular \acrshort{rl} problem is rarely addressed. In practice, the size, i.e. depth and width of the neural networks, is mainly adjusted by either random search or grid search methods~\cite{Bergstra.2012}. The other recent methods usually tune other hyperparameters  such as  learning rate, entropy cost, and intrinsic reward and do not consider size of the network in \acrshort{rl}~\cite{Jaderberg.2017}. Therefore, tuning for an optimal architecture requires knowledge on both the type of \acrshort{rl} algorithm that is used and the application domain where the algorithms are applied, which inhibits fast deployment of the learning agents. An automatic adjustment of the required network parameters is highly desirable because of the long training times in \acrshort{rl} together with the large number of hyperparameters to be tuned.

To this end, we tackle the above described weaknesses in current \acrshort{rl} methods, namely targeted training in the evolving learning  setting and the automatic adjustment of the trainable parameters in the neural network. We present \acrfull{gmrl}, which maintains trust regions and reduces gradient variance, from the initial training phase in two of those algorithms, for a targeted training. The original proposal  for  \acrfull{gm} with network pruning was originally introduced in~\cite{Chadha.2019c} for supervised training of \acrshort{dnn}. We enhance the previous work by concentrating on the gradients flow in the network rather than the weights. Specifically, rather than pruning irrelevant weights, we focus on the  adaptive learning of the most relevant weights during the course of training. We develop different methods for \acrshort{gmrl}, starting with a method that requires knowledge of the learning process and then developing a momentum based dynamic learning scheme which particularly suits the sequential learning process of \acrshort{rl}. We further develop method to automatically adjust the \acrshort{gm} hyperparameters, particularly the active network capacity required for a certain task. It is important to note that the proposed approaches are independent from the type of \acrshort{rl} algorithm used and are therefore, universally applicable. We apply and  test the proposed algorithms in various continuous and discrete application domains. The proposed \acrshort{gm} approaches with the \acrshort{a2c}~\cite{Mnih.2016} algorithm is tested on a multi-robot manufacturing station where the goal is a coordinated operation of two industrial robots, sometimes termed in literature as Job-Shop Scheduling Problem (JSSP)~ \cite{Applegate.1991}. Thereafter, we test the approach on some well known reinforcement learning environments from OpenAI Gym~\cite{GregBrockman.2016} like the Atari games from  the Arcade  Learning Environment~\cite{Bellemare.2013} and MuJoCo~\cite{E.Todorov.2012} both with the \acrshort{ppo}~\cite{Schulman.2017} algorithm. The results obtained underline the improved generalization performance and the capability to automatically adjust the network size allowing for successful training also in strongly over-parameterized neural networks.

The contributions of the work can be summarized as follows:
\begin{itemize}
    \item We introduce four novel GM methods, each successively increasing the performance of the RL algorithm, namely Frozen threshold with Gradient Monitoring (F-WGM), Unfrozen threshold with Gradient Monitoring (U-WGM), Momentum with Gradient Monitoring (M-WGM), Adaptive Momentum with Gradient Monitoring (AM-WGM).
	\item The methods reduce the gradient variance helping in improving the  training performance during the initial phase of  \acrshort{rl} with the M-WGM and AM-WGM methods acting as a replacement for  gradient clipping in PPO.  In addition to the superior  evaluation performance, the methods are shown to expedite the convergence speed by increasing the learning rates and increasing the 'k-epoch' updates in PPO.
	\item The proposed  AM-WGM method allows for continuous adjustment of the network size by dynamically varying the number of active parameters during training by adjusting  the network capacity based on the feedback from the rewards collected during the learning progress. 
	\item We conduct various experiments on different application domains including a coordination problem of a multi-robot station, Atari games, and MuJoCo tasks to underline the performance  gains and the general applicability of the proposed methods.
 \end{itemize}

The paper is organized as follows. Related work is presented in Section~\ref{sec:relwork}. In Section~\ref{sec:RL}, the basics of the \acrshort{rl}-framework employed are introduced. The proposed \acrshort{gm} methods and their integration with \acrshort{rl} is presented in Section~\ref{sec:GM}. Section \ref{sec:Results} presents a thorough comparison of the results obtained from proposed methods on the various application domains. Section~\ref{sec:conclusion} concludes the paper and an outlook for future work.


\section{Related Work} \label{sec:relwork}

We first discuss general approaches in \acrshort{dnn} training that help in better generalization capabilities and subsequently  focus on methods specifically for generalization in \acrshort{rl}. Finally, we discuss approaches for tuning the network size in \acrshort{rl}.

\textbf{Generalization in \acrshort{dnn}:} Deep feed-forward neural networks had been notoriously difficult to train in the past due to various factors including vanishing gradient~\cite{Hochreiter.1998}, highly non-convex optimization problems~\cite{Gori.1992}  and  the tendency  to over-fit~\cite{Lawrence.1997}. All of these short-comings have been virtually mitigated in modern deep learning architectures through a myriad of techniques. They include initialization of the trainable parameters~\cite{Glorot.2010,He.2015}, the use of sparse and non-saturating activation functions such as ReLU~\cite{Glorot.2011} in the hidden layers and the use of more efficient stochastic gradient descent optimization algorithms such as Adam~\cite{DiederikP.Kingma.2015}. The other approaches enhancing generalization capabilities are Batch Normalization~\cite{Ioffe.2015b} to overcome the internal co-variate shift and Dropout\cite{Srivastava.2014} which masks the neural activations with a masking matrix drawn from a Bernoulli distribution. For dropout, various variants improving on the vanilla dropout have been developed including variational dropout~\cite{Kingma.2015b} and targeted dropout~\cite{Gomez.2019}. Similarly, individual weights instead of hidden activations units are dropped in~\cite{Wan.2013}. Recently, it has been investigated that over-parameterization also leads to better generalization  performance in supervised learning with \acrshort{dnn} ~\cite{BehnamNeyshabur.2019,Belkin.2019,Brutzkus.2019}. Another popular approach is the incorporation of auxiliary loss functions into the main loss resulting in either $L_1$ or $L_2$ regularization. An increasing popular method for optimizing neural network training is gradient clipping~\cite{Pascanu.2013}, originally developed for the exploding gradient problem in recurrent neural networks. It has been proven to increase convergence speed in supervised learning in~\cite{JingzhaoZhang.2020}.  Also a multitude of approaches for network pruning have been reported to help in the generalization performance. Generally, the pruning methods are applied iteratively based on  magnitude based~\cite{Han.2015}, gradient or Hessian ~\cite{Hassibi.1993,LeCun.1990b}. Recent methods such as~\cite{NamhoonLee.2019,NamhoonLee.2020} calculate the  sensitivity of each connection and prune the weights  with a single shot approach. Please refer~\cite{Blalock.2020} for a recent overview of the various pruning methods that have been developed for neural networks. We emphasize that our approach does not include pruning weights, but freezing them by not allowing the gradients to flow to the respective weights. Also a direct application of pruning methods in \acrshort{rl} is not clear  as these methods usually require a retraining which is far-fetched  for the evolving data-set scenario   during \acrshort{rl} training. Indeed, all of the above methods that have been used in \acrshort{rl}, were specifically developed for supervised learning, but just found themselves to be used in \acrshort{rl}.

\textbf{Variance Reduction and Generalization in \acrshort{rl}:} Variance reduction techniques for gradient estimates in \acrshort{rl} have been introduced in~\cite{Greensmith.2004} where control variate are used for estimating performance gradients. An Averaged Deep Q Network approach has been proposed in \cite{Anschel.2017} where averaging previously learned Q-values estimates leads to a more stable training procedure.  Also, variance reduction in the gradient estimate for policy gradient \acrshort{rl} methods  has been proposed in~\cite{HongziMao.2019}   with an   input-dependent baseline which is a function of both the state and the entire future input sequence. Contrary to the previous approaches, we consider  variance reduction in the gradient estimate by freezing  the  gradient  update  of  a  particular  weight. 

Literature on generalization in \acrshort{rl} usually focuses on the performance of the trained agent in an unseen environment~\cite{Cobbe.2019,Justesen.2018,XingyouSong.2020,Igl.2019}. However, better generalization methods for evaluating the agent on the same environment is missing in literature. This is especially the case in industrial production environments where the production setup does not change drastically with time. The  proposed approach is focused on this area where a fast and reliable training  procedure has been developed for discrete and continuous environments.

\textbf{Neural Architecture Search:}
There are a number of hyperparameters in neural network training, with the size of the network being one of the most important ones. Apart from grid search and random search, there also exist a number of approaches including Bayesian optimization~\cite{Bergstra.2013}, evolutionary methods~\cite{Young.2015}, many-armed bandit~\cite{Li.2017b}, population based training~\cite{Jaderberg.2017} and  \acrshort{rl}~\cite{Baker.2017}. All of the above methods search for neural architectures for a supervised learning setup. \cite{Schaul.2019} present a multi-arm bandit approach for adaptive data generation to optimize a proxy of the learning progress. We on the other hand, propose a method which makes the learning process  robust to the choice of the size of the  network. Furthermore, all the of the above methods search in a sequential and very computationally expensive manner. Our proposed method on the other hand,  start with a possibly over-parameterized network and increase or decrease the learning capacity during training to adjust the learning procedure. This way we dynamically determine the actually relevant number of parameters in each training phase.


\section{Introduction to Reinforcement Learning}\label{sec:RL}

Reinforcement Learning (RL) is the branch of machine learning that deals with training agents to take an action $a$, as a response to the state of the environment at that particular time, $s_{t}$, to get a notion of reward, $r$. The objective of the RL agent is to maximize the collection of  this reward. Sutton and Barto define RL as, \say{ learning what to do – how to map situations to actions – so as to maximize a numerical reward signal} \cite{Sutton.1998}. 

\begin{figure}
    \centering
    \includegraphics[scale=0.5]{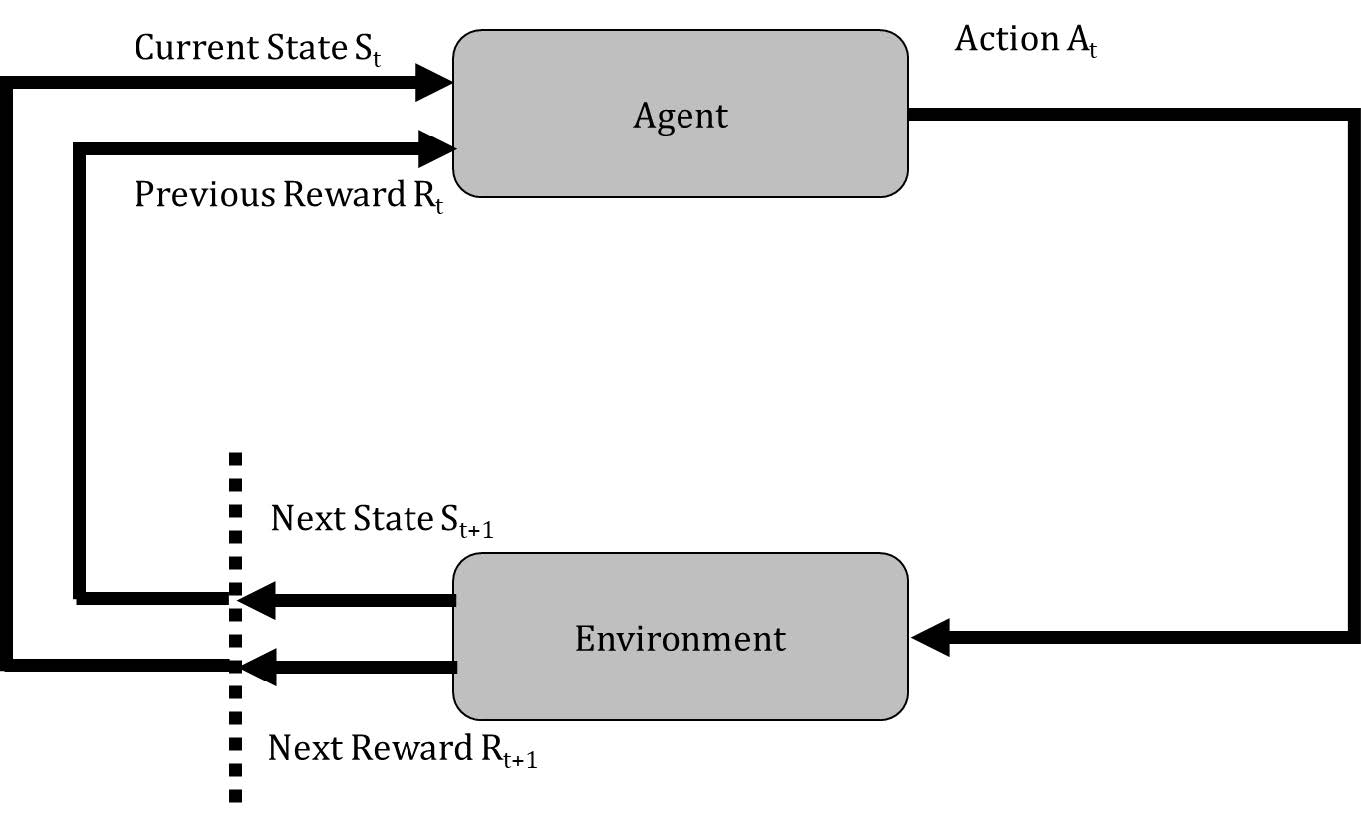}
    \caption{The interaction of agent and environment as MDP}
    \label{fig:my_label}
\end{figure} 

A reinforcement learning system has two major components: the agent and the environment where the overall system is characterized as a Markov Decision Process (MDP). The agent is the intelligent learning system, while the environment is where the agent operates.  The dynamics of the MDP is defined by the tuple $\mathcal{(S, A, P, R,}$ $\mathcal{p}_0$ $)$, with the set of states $\mathcal{S}$, the set of actions $\mathcal{A}$, a transition model $\mathcal{P}$, in which for a given state $s$ and action $a$, there exists a probability for the next state $s^{\prime} \in \mathcal{S}$ , a reward function $\mathcal{R : S \times A \times S \xrightarrow{} \mathbb{R}}$ which provides a reward for each state transition $s_t \xrightarrow{} s_{t+1}$. and a re-initialization probability $\mathcal{p}_0$.
A policy $\pi(a|s),$ provides an action $a \: \epsilon \: \mathcal{A}$, for a state $s$ presented by the environment. A policy could use \textit{state-value function}, $v(s) = E[R_{t}|S_{t}=s]$, which is the expected return from the agent when starting from a state $s$, or an \textit{action-value function}, $q(s, a)  = E[R_{t}|S_{t}=s, A_{t} = a]$, which is the expected return from the agent when starting from a state $s$, while taking action $a$. Here, $R_{t} =  \sum_{t}{\gamma^t r_{t}}$ is the discounted reward that the agent collects over $t$ times steps  and $\gamma$ is the discount factor, where $ \: 0 \leq \gamma \leq 1.$ The policy then can be defined by an $\epsilon-greedy$ strategy where the actions are chosen based on $\pi(a|s) = argmax(q(s, \mathcal{A}))$ for a greedy action or a completely random action otherwise.\\
Alternatively there are policy gradient methods that use a \textit{parameterized policy}, $\pi(a|s,\theta)$, to take action without using the value functions to take actions. Value functions may still be used to improve the learning of the policy itself as seen in A2C. The objective of the agent is to find an optimal policy, $\pi^*(a|s),$ that collects the maximum reward. To find the optimal policy, the trainable parameters of the policy are updated in such a way that it seeks to maximize the performance as defined by the cost function $J(\theta_{t})$ as illustrated in Equation \eqref{eq:policy update}. There exists at least one policy, such that $\pi^*(a|s) \geq \pi(a|s)$, where $\pi^*$ is defined as the optimal policy.
\begin{equation}
    \label{eq:policy update}
    \theta_{t+1} = \theta_{t} + \rho \nabla J(\theta_{t}),
\end{equation}
 where $\theta$ are the parameters of the policy $\pi$ and $\rho$ is the learning rate.  There are different choices of $J(\theta_{t})$ for different algorithms as explained in the sections below.

\subsection{Advantage Actor-Critic}
In this section we introduce the policy gradient algorithm Advantage Actor Critic (A2C)~\cite{Mnih.2016}. A2C are policy gradient methods that use the value function to reduce the variance in the calculated cost function. Here, the \textit{actor} refers to the policy $\pi (a|s,\theta_{1})$ and the \textit{critic} refers to the value function $v(s,\theta_{2})$, where $\theta_{1} \: and \: \theta_{2} $ are the parameters of the actor and critic respectively. The parameters $\theta_{1} \: and \: \theta_{2} $ are partially shared in case of the A2C algorithm we use. The cost function for the actor and the critic of A2C algorithm is given by Eqn. \eqref{a2c-j} and \eqref{adv} respectively.

\begin{equation}
    \label{a2c-j}
    J(\theta) = \mathbb{E}_{\sim \pi_{\theta}} \left[ \sum_{(s_{t},a_{t}) \epsilon } \: log \pi_{\theta}(a_{t}, s_{t}) \: . \: A_{\pi_{\theta}} (s_{t}, a_{t}) \right] 
\end{equation}

\begin{equation}
    \label{adv}
    A_{\pi_{\theta}} (s_{t}, a_{t}) = Q_{\pi_{\theta}} (s_{t}, a_{t}) - V_{\pi_{\theta}} (s_{t})
\end{equation}

We use two co-operative agents that use indirect communication channels to solve the multi robot coordination environment. They are explained in detail in section \ref{learn_setting}.

\subsection{Proximal Policy Optimization:}

In this section, Proximal Policy Optimization (PPO) is explained. In A2C, the focus of the algorithm was to get a good estimate of the gradients of the parameter $\theta$. But applying multiple optimization steps on this, empirically leads to large policy updates that destabilizes the learning process. A surrogate objective function in used in PPO to overcome this. 

\begin{equation}
    \label{ppo}
    \underset{\theta}{max} \: \mathbb{E}_{\sim \pi_{\theta}} \left[ min ( r_{t}(\theta)\hat A_{t}, \: clip(r_{t}(\theta), 1-\epsilon, 1+\epsilon) \hat A_{t})  \right]
\end{equation}
where
\begin{equation}
    r_{t}(\theta) = \frac{\pi_{\theta}(s_{t}, a_{t})}{\pi_{\theta_{old}}(s_{t}, a_{t})}
\end{equation}

and $\hat{A_{t}}$ is the estimator of the advantage function at time step $t$. Refer to  ~\cite{Schulman.2017} to have a full overview of the algorithm. Due to this controlled nature of the policy updates PPO are found to work well with continuous control problems. Hence, PPO is used in MuJoCo and Atari Learning Environment. 

\section{Reinforcement Learning with Gradient Monitoring}\label{sec:GM}

Modern deep learning architectures are in general over-parameterized, i.e. the parameters drastically outnumber the available data set size~\cite{Arora.2018b,BehnamNeyshabur.2019}. While this has been empirically shown to improve the learning performance compared to shallow architectures, determining a suitable number of layers and neurons depending on the problem at hand remains to be an open issue. 
To circumvent the determination of network size, successful workarounds have focused on reducing the number of actively learning parameters per iteration. They end up in reducing the degrees of freedom during the training using methods like drop-out, drop connect and their various sub-forms where network activations or weights are randomly switched off.\\
Gradient monitoring follows a different approach in that it  intends to steer the learning process of the \acrshort{dnn} by actively manipulating the backward pass of the training process. Specifically, we purposefully  deactivate and activate the gradients in the backward pass for a subset of weights based on the learning conditions which is explained in the subsection below. Although applicable to deep learning setting, we find GM particular useful for reinforcement learning since it reduces the variance in the gradient estimates during the crucial initial part of the learning process and also introduces a dynamic way to clip the gradients that is applied layer-wise as opposed to the norm of the entire gradients popularly used.

\subsection{Gradient Monitoring in DNN}

To illustrate the training procedure with GM, we consider fully connected feed-forward DNN with more than one hidden layer trained with mini-batch gradient descent and gradient based optimizers, although we found it most effective with momentum based gradient optimizers like Adam~\cite{DiederikP.Kingma.2015}. However, we emphasize that GM is universally applicable to other network structures like convolutional or recurrent NN. The training procedure in NN minimizes the loss function by calculating the partial derivative of the loss functions with respect to each of the weight parameters recursively. Hence, for a NN model with $m \geq 2$ hidden layers we denote $W_{m}$ as the weight matrix for the $m^{th}$ layer and $\nabla L_{W_1}$, $\nabla  L_{W_2} $ .... $ \nabla  L_{W_m}$ denote the gradients for each weight matrix. The gradient calculated as per the Adam optimizer is shown in Equation \eqref{eq:weightupdate}
\begin{equation}
    \label{eq:weightupdate}
    \nabla L_{W_t}=\frac {\hat{{m_t}}}{{\sqrt{\hat{{v_t}}}+\epsilon}}.
\end{equation}
To deactivate the gradients, we  set elements of the gradient matrix $\nabla L_{W_t}$ in~\eqref{eq:weightupdate} to zero. To accomplish this, we define a masking matrix $M$, whose values are either one or zero, and calculate the new gradient matrix $ \nabla \hat {L}_{W_t}$ as shown in ~\eqref{mask_mul}.
\begin{equation}
    \label{mask_mul}
	\nabla \hat{L}_{W_t} = M_{W_t} \circ \nabla L_{W_t},
\end{equation}
 where $\circ$ denotes the Hadamard product. The weight update is performed then with a standard gradient descent update as in Equation~\eqref{eq:weight_step}
\begin{equation}\label{eq:weight_step}
    {W}_{t+1} = {W}_t - \rho \nabla \hat L_{W_t}.
\end{equation}

The steering of the learning process is decided based on the effect of each parameter on the forward as well as the backward pass. Therefore, the masking matrix, $M_{W_t}$, is calculated based on a function that takes as input, the weights $W_t$, their respective gradients $\nabla L_{W_t}$ from the backward pass, a learning threshold $\mu(W_t, \nabla L_{W_t})$, and a learning factor $\lambda$. A decision matrix $D_{W_t}(W_t, \nabla L_{W_t})$  is constructed to estimate the learning process. This decision matrix $D_{W_t}(W_t, \nabla L_{W_t} )$ is compared with the  learning threshold  $\lambda \mu (W_t, \nabla L_{W_t})$, in order to make the decision if the masking value is active (1) or inactive (0). The decision matrix can be calculated using lot of combinations like $\left|\frac{\nabla L_{W_t}}{W_t}\right|$, $\left|\frac{{W_t}}{\nabla L_{W_t}}\right|$ or $\left|\nabla L_{W_t} \circ {W_t} \right|$. We use the absolute values since we are interested in the quantum of learning. Specifically, the masking matrix $M$ can be defined as
\begin{equation}
	M_{W_{t}} =  {H} (D_{W_t}(W_t, \nabla L_{W_t}) - \lambda \mu(W_t, \nabla L_{W_t})),
\end{equation}
where ${H}$ is the Heaviside step function in which the gradients of weight connections which do not reach the relative amount of learning are deactivated, i.e. receive no gradient during the back-pass. Note that due to the use of Adam optimizer, the decision for freezing gradients is not only based on the actual gradient calculated over a mini-batch but based on the decaying average of the previous gradients. We emphasize that GM is applied to each layer in the NN. The list of hyperparameters used along with their symbols and the algorithm they are used in is given in table \ref{tab:hyperparameters}

\begin{table}
\caption{Hyperparameters used in GM algorithms}

\begin{center}
\begin{tabular}{ |c|c|c|c|c| } 
\hline
Symbol & Description & VGM & M-WGM & AM-WGM\\
\hline
 $\lambda$ & Learning factor& \cmark& \cmark& \cmark\\
 $\eta_{start}$ & Start of GM & \cmark & \xmark& \cmark\\
 $\eta_{repeat}$ & Mask update frequency& \cmark & \xmark& \cmark\\
 $\zeta$ & Masking momentum& \xmark& \cmark& \cmark\\
 $M_\zeta$ & Momentum matrix& \xmark& \cmark& \cmark\\
 $\alpha_\lambda$ & change rate of $\lambda$& \xmark& \xmark& \cmark\\
 $\phi$ & Reward collection rate& \xmark& \xmark& \cmark\\
 $R$ & Rewards collected& \xmark& \xmark& \cmark\\
\hline
\end{tabular}
\label{tab:hyperparameters}
\end{center}
\end{table}

\subsection{Vanilla Gradient Monitoring}

The core of GM is the derivation of suitable conditions for activating and deactivating the gradients, $\nabla L_{t}$, flow which includes deriving $\mu$ based on the actual status of learning. To keep the representation simple $D_t(W_t, \nabla L_{W_t})$ and $\mu(W_t, \nabla L_{W_t})$ will be simply written as $D_t$ and $\mu$ respectively henceforth. Obviously, keeping a constant integer value as the learning threshold $\mu$ for all the gradients is not appropriate as the proportion of learning represented in the gradients might have different distributions in different layers and different time-steps. Furthermore, choosing a single constant learning value for different learning  tasks is not trivial. Hence, the learning threshold is made adaptable by the use of functions like the mean or the percentile of the values of the decision matrix $D_{W}$. This provides a method that ensures that a certain portion of the gradients are allowed in any situation. We define $H$ such that all gradients above or below the learning condition is deactivated. In this paper we use the mean of all the elements $d_{ij}$ in the decision matrix $D_{W_t} \in \mathbb{R}^{n}$ for each layer $m$ as the $\mu$ function, and use $\left|\frac{\nabla L_{W_t}}{W_t}\right|$ as the $D$ function. Concretely, we deactivate all gradients below this learning condition:

\begin{align}
    \mu_{m} = \frac{1}{n} \sum_{ij} d_{ij}
\end{align}

Beside the question which gradients to deactivate, we also have to answer the question when to deactivate the ineffective gradients to make training most effective. This problem is solved in two ways. First, as with the learning rate, similar schedules for deactivating is set up depending on the problem at hand. The methods F-WGM and U-WGM use this setup which are together called Vanilla Gradient Monitoring (VGM).\\
Alternatively, we introduce a momentum parameter on top of the masking matrix to alleviate the problem in deciding when to start deactivating the gradients. The methods M-WGM and AM-WGM use these methods. In this section, we further discuss only about the methods F-WGM and U-WGM, while M-WGM and AM-WGM are discussed in the further sections.\\
For F-WGM and U-WGM we have to define $\eta_{start}$, which defines after which epoch the masking matrix is applied, along with the $\lambda$ parameter, which is the multiplication factor for the learning condition $\mu$. $\eta_{start}$ is a hyperparameter which is tuned. But the start of the GM application can be automated by giving it as the point of the first successful episode. This is the point near which where we found empirically the highest amount of gradients being back propagated. So creating and applying the first masking at this point makes sense. The pseudo code for F-WGM and U-WGM is provided in \ref{alg:gm}. The only difference between F-WGM and U-WGM is that, in the case of F-WGM the $\lambda$ is kept constant and $M$ is updated with the same $\lambda$ value for every few iterations ($\eta_{repeat}$). While in U-WGM, the $\lambda$ is made variable, decreasing in value after every update

\makeatletter
\def\BState{\State\hskip-\ALG@thistlm}
\makeatother

\begin{algorithm}[tb]
	\caption{Frozen and Unfrozen with Gradient Monitoring}
	\label{alg:gm}
	\begin{algorithmic}[1]
		\State {\bfseries Input:}  $\nabla L_{t}$, \textit{${W}_{t-1}$}, $\rho$, $\lambda$, $\eta$, $\eta_{start}$, $\eta_{repeat}$
		\State \textbf{Init:} Masking Matrix ${M}$
		\State \textbf{Sequence:} 
        \If{$\eta >= \eta_{start}$} and $\eta_{repeat} \% \eta == 0$
            \For{each layer $m$}
    		\State Masking matrix $M={H} \big(D_t -  \lambda \mu$ \big)  		
    		\State Gradients: $\nabla L_{t} = \nabla L_{t} \circ {M}$
    		\EndFor
		\EndIf
		\State \textbf{Output:} Weights \textit{${W}_{t}$} = \textit{${W}_{t-1}$} + $\rho\nabla L_{t}$ 
	\end{algorithmic}
\end{algorithm}

The motivation behind the U-WGM is that the weight parameters which did not have a relative high impact on the learning process during the initial phase of learning (till epoch $\eta_{start}$) might nevertheless have an impact later, e.g. once other weights have settled. Hence, by reducing the learning condition threshold, those weights can participate in the learning process again. The factor $\lambda$ is a hyperparameter which in practice we found that halving, i.e. $\lambda^\prime=\lambda/2$ at every $\eta_{repeat}$ works well.

\subsection{Momentum with GM}

One of the disadvantages of the previous approaches is that the performance of the algorithm is hugely dependant on the hyperparameters $\eta_{start}$ and $\eta_{repeat}$. $\eta_{start}$ is at the first episode of convergence, since that was around where the absolute sum of gradients was at the maximum. This poses a problem when scaling up the use of GM RL to other continuous control long-horizon tasks, since we always need to decide in hindsight when to begin the application of GM. Hence a new version of GM RL was developed to tackle the same called Momentum with Gradient Monitoring (M-WGM). Here we introduce a momentum matrix $M_{\zeta}$ and a momentum hyperparameter $\zeta$, where the momentum matrix applied to the gradients right from the first episode and the momentum hyperparameter provides an intuitive control over the learning process. The pseudo code for M-WGM is give in Algorithm \ref{alg:mgm}.

\begin{algorithm}[tb]
\caption{Momentum - Gradient Monitoring}
	\label{alg:mgm}
	\begin{algorithmic}[1]
		\State {\bfseries Input:}  $\nabla L_{t}$, \textit{${W}_{t-1}$}, $\rho$, $\lambda$, $\zeta$
		\State \textbf{Init:} ${M_{\zeta}}$, ${M}$
		\State \textbf{Sequence:} 
		\For{each layer}
		\State Masking matrix $M={H} \big(D_t -  \lambda \mu$ \big)
		\State Momentum matrix: $M_{\zeta} = M_{\zeta}{\zeta}  +  M(1-{\zeta})$
		\State Gradients: $\nabla L_{t} = \nabla L_{t} \circ M_{\zeta}$
        \EndFor
		\State \textbf{Output:} Weight \textit{${W}_{t}$} = \textit{${W}_{t-1}$} + $\rho\nabla L_{t}$ 

	\end{algorithmic}
\end{algorithm}

The gradients and the masking matrix are calculated as usual, but the masking matrix $M$ is not used directly. We use the momentum matrix $M_\zeta$ which now keeps track of the running momentum of the elements of the masking matrix. The momentum matrix is element-wise multiplied with the gradients and the gradients are finally applied to the weights. The rationale behind this being that the gradients are updated according to the frequency of their activation in the masking matrix. So instead of applying the continuously varying noisy masking matrix, we instead use a controlled momentum matrix. The momentum method controls the variance in the gradient updates, especially in the early stages of the RL learning process where this stability provides for convergence to a better minima. Also in the later stages the momentum matrix still controls the sudden bursts of gradients that could destabilize the learning process and therefore provides much better performance of the agents as shown empirically in the results section. As such we use this method as a controlled replacement for the global gradient clipping usually done in RL algorithms.

\subsection{Adaptive Momentum with GM}

The $\lambda$ parameter in the M-WGM algorithm is kept constant throughout the training. But as noticed in the U-WGM method,  modifying the hyperparameter for learning condition threshold ($\lambda$) improves performance. Hence in this section, we introduce the algorithm Adaptive Momentum with Gradient Monitoring (AM-WGM), where instead of hand setting the threshold for masking matrix activation it is made adaptable based on the performance (reward collection rate $\phi$) of the agent. For example, if the agent performs worse than before then the threshold is increased so that fewer gradients are active in the masking matrix and vice-versa. This means when the performance is bad, learning is restricted to fewer weights, while in case of improved performance, the agent is provided more weights in the learning process. The pseudo code for the AM-WGM is provided in Algorithm \ref{alg:amgm}. To ensure stability in the initial training episodes, $\lambda$ is not modified until a few episodes are completed, usually at about 30\% of the total episodes, and it is also updated after every few episodes, usually at about 10\% of the total episodes. These are also denoted by the hyperparameters $\eta_{start}$, $\eta_{repeat}$. 

\begin{algorithm}[tb]
\caption{Adaptive Momentum with Gradient Monitoring}
	\label{alg:amgm}
	\begin{algorithmic}[1]
		\State {\bfseries Input:} $\nabla L_{t}$, \textit{${W}_{t-1}$}, $\rho$, $\lambda$, $\zeta$, ${\alpha_\zeta}$, $\eta$, ${\eta_{start}}$, ${\eta_{repeat}}$

		\State \textbf{Init:} $M_\zeta$, $M$, ${R_{o}}$, ${\phi_{n}}$, ${\phi_{o}}$, ${\eta_{start}}$, ${\eta_{repeat}}$
		
		\State \textbf{Sequence:}
        \State Update ${R_{n}}$
		\If{${\eta} >= {\eta_{start}}$ and $\eta \% \eta_{repeat}$ == 0}
		    \State ${\phi_{n}}$ = ${R_{n}}$ $/$ ${R_{o}}$ 
		    \If {${\phi_{n}}$ $/$ ${\phi_{o}}$ $>=1$} 
		        \State \textit{change} = -1
		    \ElsIf {${\phi_{n}}$ $/$ ${\phi_{o}}$ $<1$}
		        \State \textit{change} = 1
		    \EndIf
		    \State $\lambda$ = clamp($\lambda$+(${\alpha_{\lambda}}$*\textit{change}), 0, 1)
		    \State ${\phi_{o}}$ = ${\phi_{n}}$
		    \State ${R_{o}}$ = ${R_{n}}$
		\EndIf
		
	    \For{each layer}
		\State Masking matrix $M={H} \big(D_t -  \lambda\mu \big)$
		\State Momentum matrix $M_\zeta = M_\zeta{\zeta}  +  M(1-{\zeta})$
		\State New Gradients: $\nabla L_{t} = \nabla L_{t} \circ M_{\zeta}$
        \EndFor
        \State \textbf{Output:} Weight \textit{${W}_{t}$} = \textit{${W}_{t-1}$} + $\rho\nabla L_{t}$

	\end{algorithmic}
\end{algorithm}

\begin{align}
    R_n = \frac{1}{T} \sum_{i=1}^{T} r_t
\end{align}

So AM-WGM is similar to M-WGM in the initial stages of the learning process and it only activates after a certain number of updates are applied and the learning process has stabilized. The algorithm initializes the parameters: reward collected in current episode ($R_n$), reward collected in previous episode ($R_o$), rate of reward collected in the current episode ($\phi_n$), rate of reward collected in the previous episode ($\phi_o$). The mean reward collected is stored for the current episode is stored as $R_n$. The rate of reward collected, $\phi_n$, is the calculated. If the rate of reward collection increased ($\phi_n$ $/$ $\phi_o$ >= 1) then we reduce the threshold ($\lambda$) (which allows more gradients to be used), while we increase the increase the threshold ($\lambda$) if the performance has degraded. The hyperparameter $\alpha_\lambda$ controls the amount of change in the $\lambda$ value. The adaptable nature of the algorithm has empirically shown to increase the performance of the agents.

\subsection{Summary:}
The GM methods explained above contribute mainly on two things on the algorithm level: provide an additional trust-region constraint for the policy updates and variance reduction of the gradients. The additional trust-region constraint is provided by removing the noisy insignificant gradient contributions. Noise in the gradients update is reduced by use of the masking matrix or the momentum matrix while the insignificant contributions are removed by using the Heaviside step function. So only the consistently high contributing gradients are propagated while the others are factored to have a low impact on the learning process. The removal of these gradients also reduce the overall variance. This is especially critical for a stable learning in the initial stages of the learning process. Our results from the experiments corroborate this.
\section{Experimental Results}\label{sec:Results}

We test the GM-RL algorithms on a variety of different environments to prove its applicability empirically. We discuss and  apply the proposed methods to a real world multi-robot coordination environment with discrete state and action space. Further we apply two algorithms, M-WGM and AM-WGM, on the OpenAI Gym environments of Atari games (continuous state space and discrete action space) and MuJoCo simulations (continuous state space and continuous action space). This is because the algorithms, M-WGM and AM-WGM, perform the best in multi-robot co-ordination problem and they can also be directly applied without any hindsight information. The results from the OpenAI Gym environments prove the general 'plug and play' nature of the M-WGM and AM-WGM methods, wherein any previously proven algorithm can be improved upon by usage of GM-RL. All the \acrshort{rl} and proposed GM methods have been developed using  PyTorch~\cite{Paszke.2019b}.\\
The results section is structured as follows. All the GM methods introduced in the previous section (F-WGM, U-WGM, M-WGM, AM-WGM) are tested on the multi-robot coordination environment. The results show that the algorithm gets progressively better with each iteration. Then the applicable GM solutions (M-WGM, AM-WGM) are applied in the OpenAI Gym environments. The results obtained are compared with similar algorithms Without Gradient Monitoring (WOGM). The results are tested on various random seed initialization (Atari: 5 seeds and MuJoCo: 10 seeds) to test for stability of the algorithm. 
\subsection{Multi-Robot Coordination Environment}

This section describes the application of GM-RL algorithm on a cooperative, self-learning robotic manufacturing cell. The environment along with the corresponding RL agent setup is described in sections below, followed by the results achieved on the various trials.

\subsubsection{Environment description}

We use a simulation of the cooperative, self-learning robotic manufacturing cell to interact with the RL agent. Training is done on the simulated environment since training the RL agent on the real manufacturing cell is  time consuming and it requires large amounts of data to converge to a good minima~\cite{zhu2017target}. The simulated environment closely resembles the actual environment, emulating all the necessary features like position of work piece, status of machines etc., accelerating the learning process from taking weeks to few hours. The simulation environment is developed in Python.
\subsubsection{Learning Set-up} \label{learn_setting}

\begin{figure}[t]
    \centering
    \includegraphics[scale=0.3, align=c]{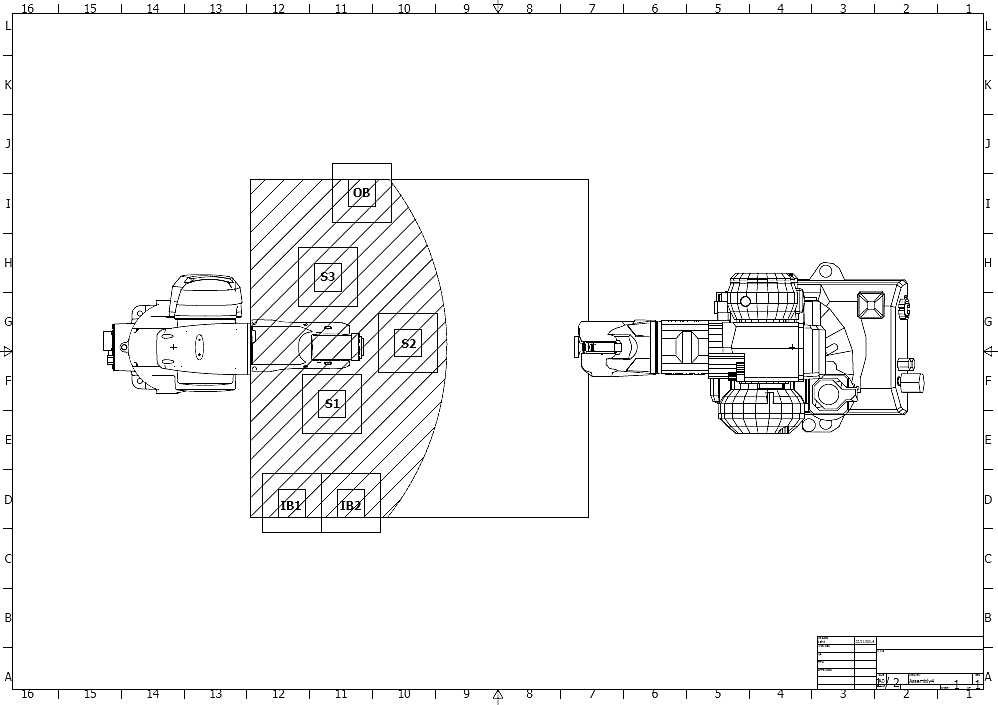}
    \caption{Schematic diagram of the multi-robot setup with the common operation region of both robots shown in grey stripes}
    \label{fig:robot_scheme}
\end{figure}

\begin{figure}[t]
    \centering
    \includegraphics[scale=0.8, align=c]{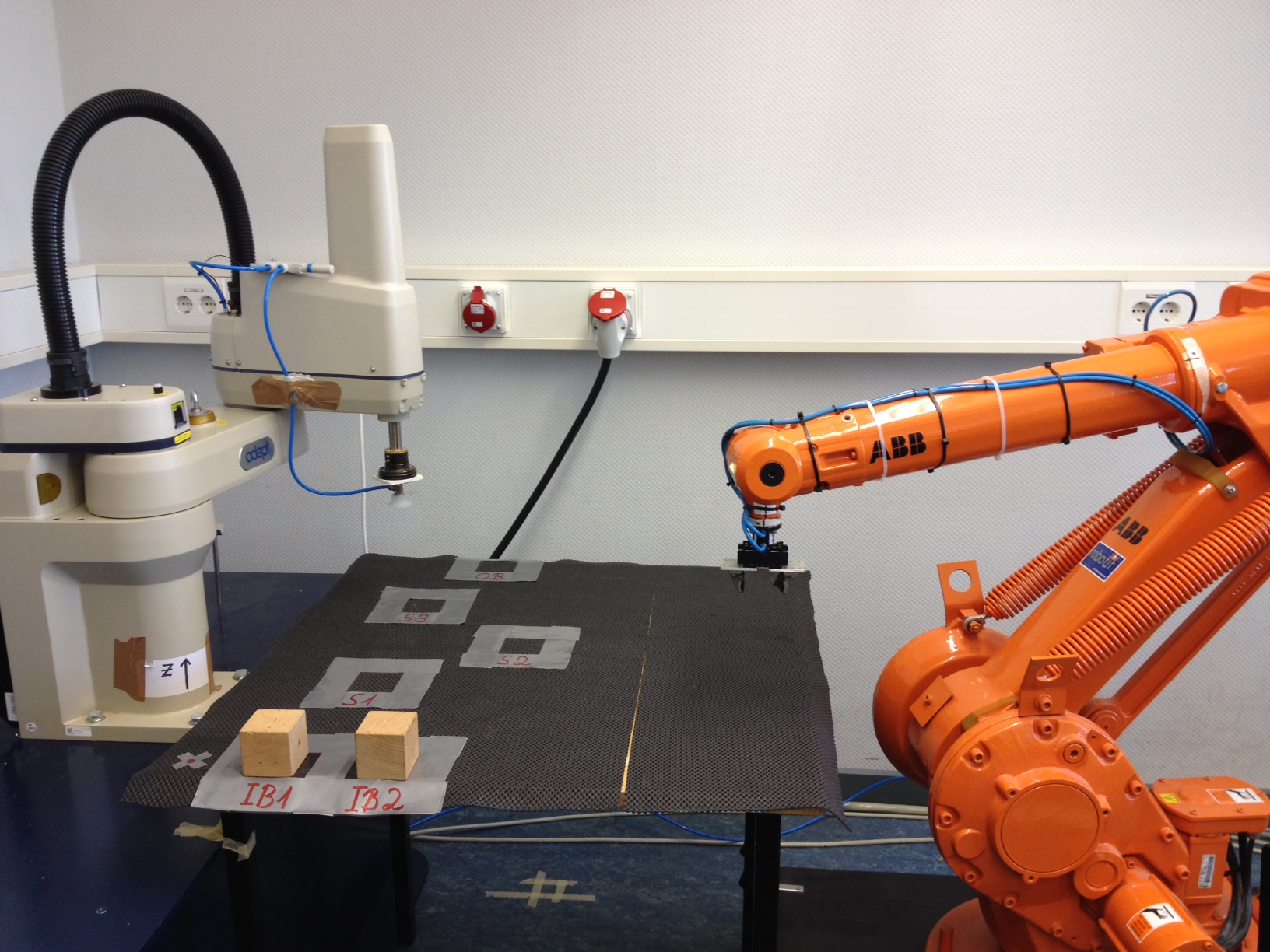}
    \caption{The multi-robot setup of two industrial robots and a working platform}
    \label{fig:robot_setup}
\end{figure}

The multi-robot coordination problem is setup in the cooperative test-bed as shown in the Figure~\ref{fig:robot_setup}. The test-bed has a dual robot setup, consisting of Adept Cobra i600 SCARA robot and an ABB IRB1400 6-DOF-robot. The test-bed has six handling stations, two input buffers, three handling stations, and one output buffer as shown in Figure~\ref{fig:robot_scheme}. There are two different types of work-pieces to be handled through the stations, Work-Piece-1 (WP1) and Work-Piece-2 (WP2), each with their respective input buffers. Both the work pieces have their own pre-defined sequences that are encoded into them through embedded  RFID chips. The schematic diagram of the test-bed is given in the Figure~\ref{fig:robot_scheme}. The robots pick and place the work-pieces from one station to the other. The work space where the handling stations are located is accessible to both the robots, hence it is possible to have a shared working space denoted by the striped grey area in Figure~\ref{fig:robot_scheme}. Each robot has its own proprietary software to control its movements. Hence a supervisory control system is implemented through Siemens S7 platform that controls and coordinates the robot movements. This supervisory control system sends signals to the robot's on-board control system where the actual action is taken by the robot. The task of the agents in this test-bed is to move a predefined number of work pieces through the handling  stations within an optimal steps or time-frame into the output buffer. 

\textbf{Agent, Environment and Reward Representation:}
In this part, we discuss the agent-type, number of agents, the action-space and the state representation for the agents. The \textit{robots} are used as the agents, where both robots act independently in a multi-agent setup. The robots take the universal state of the system and output an action each.  For the \textit{architecture} of the robotic multi-agent, independent learners with communication enabled between the agents was chosen since it is established that communication between the agents helps in faster convergence~\cite{Panait.2005}. This setup gives the RL-agent a good overview of the global conditions and local conditions. For the action-space, the agent has only predefined movements of the work-piece paths since a supervisory control system takes care of the hardware level movements. Providing this information, instead of the agent computationally finding the best movements is a practical choice as well because such a constraint already exists as a part of the manufacturing document in the factory. The action-space by extension controls the input, output, and loading of the work-piece in the resources. Additionally, no Programmable Logic Control (PLC) programming is required to implement the RL-agent. This eliminates ineffective actions and the size of the action-space is reduced to 10 instances per robot.
The state-space is a piece of very important information since this is what the agent sees. The state of the resources was chosen to represent the state-space since it was independent of the work-order size, and also had the computational advantage of being in a discrete space. The state-space has 27 states given by $3^3$, three work stations and three job types (WP1, WP2, and empty). Additionally the work-piece completion percentage is also given as part of the state-space. This acts as a communication channel between the agents to identify the work done by the other agent.

\begin{table}
\caption{Rewards setup for multi-robot environment}

\begin{center}
\begin{tabular}{ |c|c| } 
\hline
State & Reward\\
\hline
 Each step & -1\\
 Locked state & -100\\
 Incomplete after 1000 steps & -100\\
 Unequal WP movement & -30\\
 WP Output & +50\\
 Target achieved & 500\\
\hline
\end{tabular}
\label{tab:jssp-rew}
\end{center}
\end{table}

The setting-up of the \textit{rewards} for a RL-agent is important as it can influence the stability of the system. We setup the reward as follows. Every action taken by the robots incurred a reward of -0.1. During the course of the study of environment two states were identified as being locked, meaning no further movement of the job is possible. If the robot reached these states it gets a reward of -100. Also, if the agent is not able to reach the required target in 1000 steps, it receives a reward of -100. To ensure that the equal quantities of WP1 and WP2 are processed, a constraint was placed on the system such that if one of the work-piece reaches completion without the other work-piece reaching even 75\% of its target, then it gets a reward of -30 as this behaviour is not completely bad but something to be improved upon. Every individual output from the environment incurred a reward of +50, while the agent gets a reward of +500 if the global targets of both agents are achieved. The reward for the individual output can be characterised as an intermediate reward, which guides the agent to make more such actions that will eventually lead to achieving the global target. The global target is set as 20 work-pieces each of WP1 and WP2. The rewards are shown in table \ref{tab:jssp-rew}

We use a similar approach as presented in \cite{Mnih.2016} with neural network architecture in the actor critic algorithm. The main idea is to use multi-task learning \cite{Caruana.1997}, which constructs neural networks that enable generalised learning of the tasks, in the context of the actor-critic algorithm. The ‘multi-headed neural network’ also is known to generalise the tasks by taking advantage of the system specific information from the signals \cite{Caruana.1997}. The primary hyper-parameters in focus here are the network size, the learning rate, the batch size and the $n$-step size. The values of hyper-parameters which gave the best results are shown in Table \ref{tab:hpy-a2c} which were set using  grid-search. Although the network size can be arbitrarily large, we use this particular size which gave the best result for the WOGM algorithm. This is discussed in \textit{Robustness to Choice of Network Size} of the results section. The activation function in the actor and critic layer is ReLU while those used in the shared layer is Sigmoid. 

\begin{table}[]
\caption{Hyperparameters of A2C algorithm}
\begin{center}
\begin{tabular}{ |c|c|c| } 
\hline
Algorithm & Hyperparameter & Value\\
\hline
 All & NN Input Size & 29\\
 All & Body - Network Layers & 2\\
 All & Body - Layer Neurons & 10\\
 All & Head - Network Layers & 2\\
 All & Head - Layer Neurons & 10\\
 All & Batch size    & 10\\
 WOGM & Learning rate ($\rho$) & 1e-3\\
 VGM, M-WGM, AM-WGM & Learning rate ($\rho$) & 2e-3\\
 All & Learning Factor ($\lambda$) & 0.5\\
 All & Discount factor ($\gamma$) & 0.99\\
 AM-WGM & Momentum value ($\zeta$) & 0.0005\\
 AM-WGM & AM-WGM start ($\eta_{start}$) & 1500\\
 AM-WGM & AM-WGM repeat ($\eta_{repeat}$) & 1000\\
 AM-WGM & Masking Momentum ($\zeta$) & 0.999\\
 AM-WGM & Threshold change ($\alpha_\zeta$) & 0.001\\
\hline
\end{tabular}
\label{tab:hpy-a2c}
\end{center}
\end{table}

\subsubsection{Results}
\begin{figure}[t]
    \centering
    \includegraphics[scale=0.2, align=c]{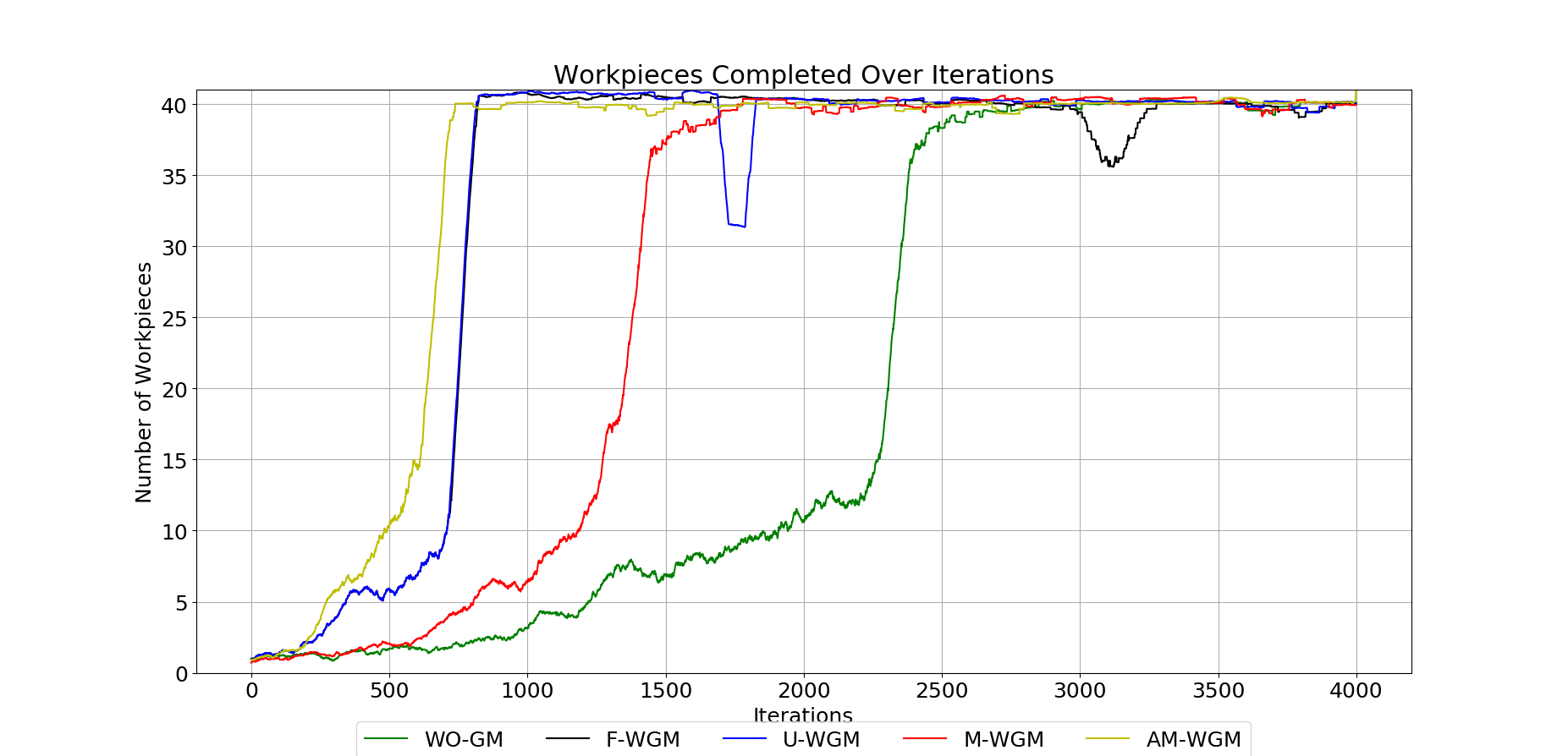}
    \caption{Convergence speed of algorithms}
    \label{fig:jssp_conv}
\end{figure}

In this section, we discuss the results in three sub-parts namely the  gradients during the back-propagation,  the amount of rewards the agents collected, and the task time and the number of work-piece outputs achieved by each agent in the multi-robot environment. All the results provided here are from the deployment of the agents after the training is complete, although we also notice that the GM-RL algorithms improve the training performance, leading to faster convergence in all GM methods as shown in Figure~\ref{fig:jssp_conv}. All the agents are trained for 5000 episodes where the convergence is achieved after 1000-2000 episodes in each of the algorithms, allowing for 3000 further episodes of learning. The agents are eventually tested for 4000 episodes.

\begin{figure}[t]
    \centering
    \includegraphics[scale=0.2, align=c]{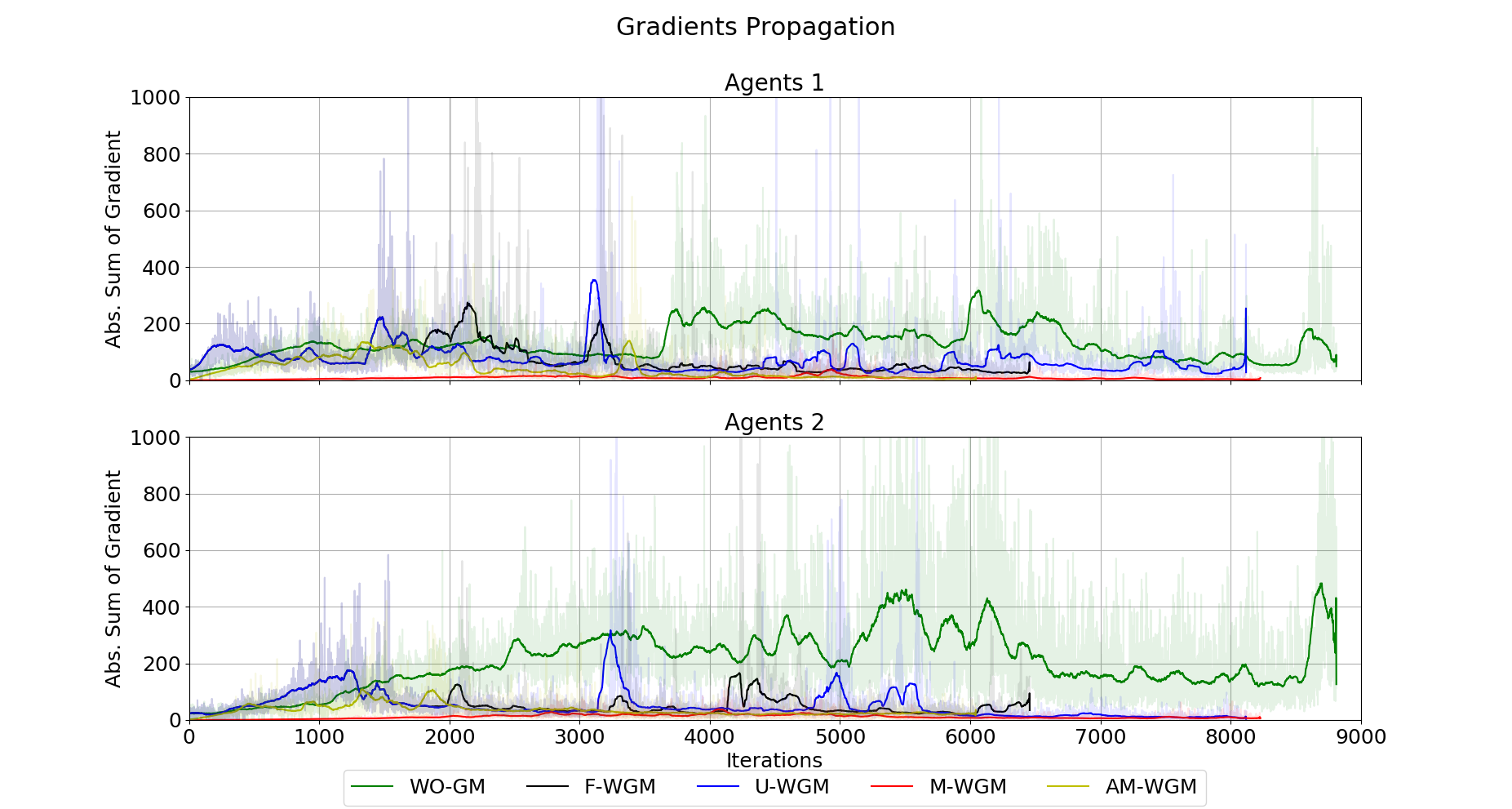}
    \caption{Gradients Propagating through the network}
    \label{fig:abs_grads}
\end{figure}

\begin{figure}[t]
    \centering
    \includegraphics[scale=0.2, align=c]{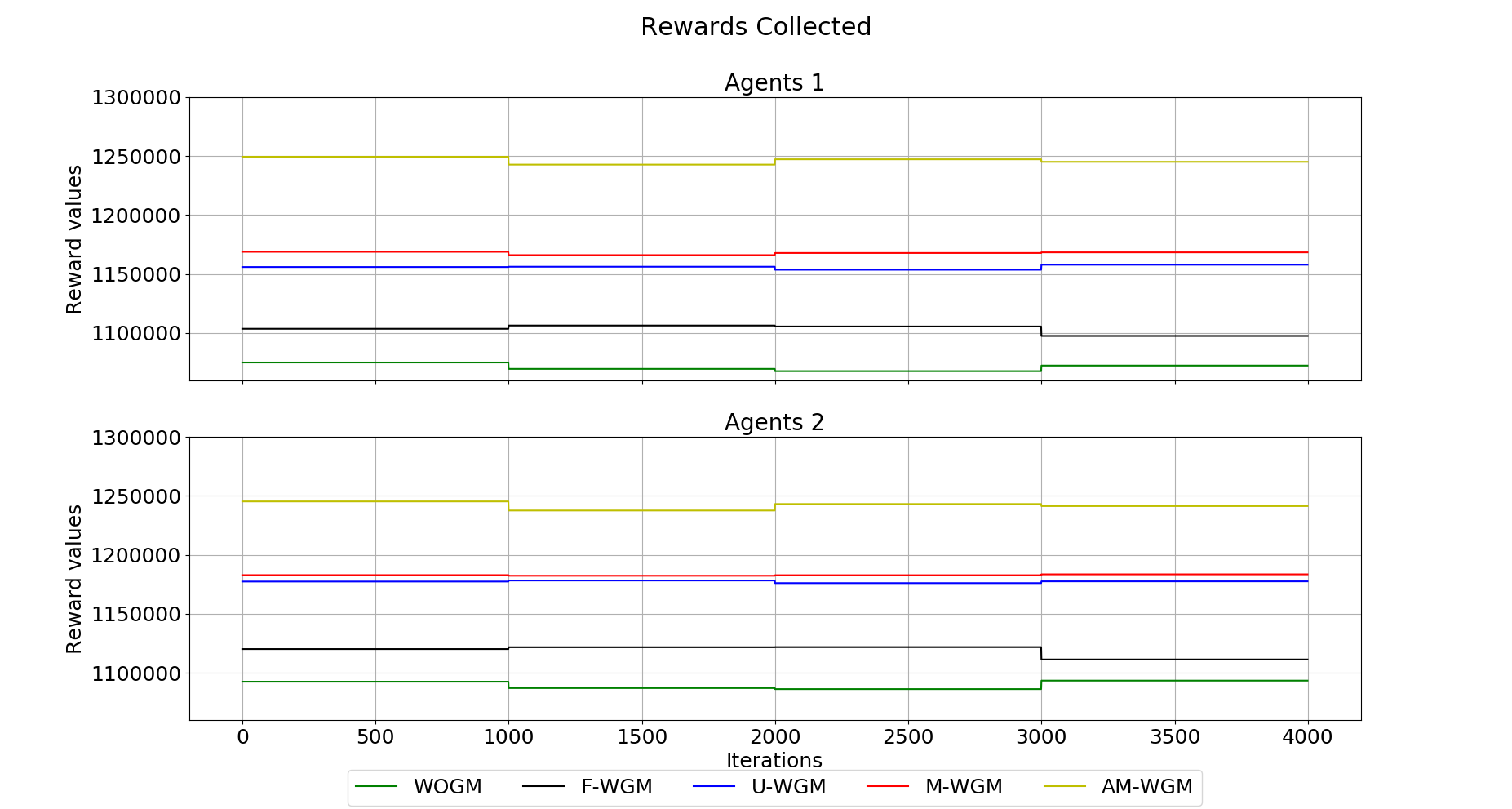}
    \caption{Rewards collected by each algorithm}
    \label{fig:rewards}
\end{figure}

\textbf{Gradients:}
In this section, we discuss about the amount of gradients that back-propagate through the neural network to analyse the targeted learning activity in the network. Since the gradient directions can be both positive and negative, in-order to get the actual quantum of the gradients, the absolute sum of the gradients for each backward pass is calculated. The absolute sum for the GM methods are calculated after the masking matrix is applied hence the quantum of gradients back-propagated by the GM methods are considerably less than the WOGM method as can be seen in Figure~\ref{fig:abs_grads}. It can be noted that in the case of U-WGM the gradients spike is noted in the iterations at which the masking matrix is applied. While the F-WGM and U-WGM are still prone to the odd fluctuations in the gradients back-propagated, it should be noted that the momentum based GM methods (M-WGM and AM-WGM) control their gradient variations well during the entire training process. The WOGM training is exposed to extensive variation in the amount of gradient flow. This variance  reduction eventually leads  to a stable learning process which reflects in the rewards collected as well as illustrated in Figure ~\ref{fig:rewards}. The AM-WGM algorithm collects the most rewards, followed by the rest of the rest of the GM methods, with the WOGM algorithm collecting the least amount of rewards.

\begin{figure}[t]
    \centering
    \includegraphics[width=0.5\textwidth, align=c]{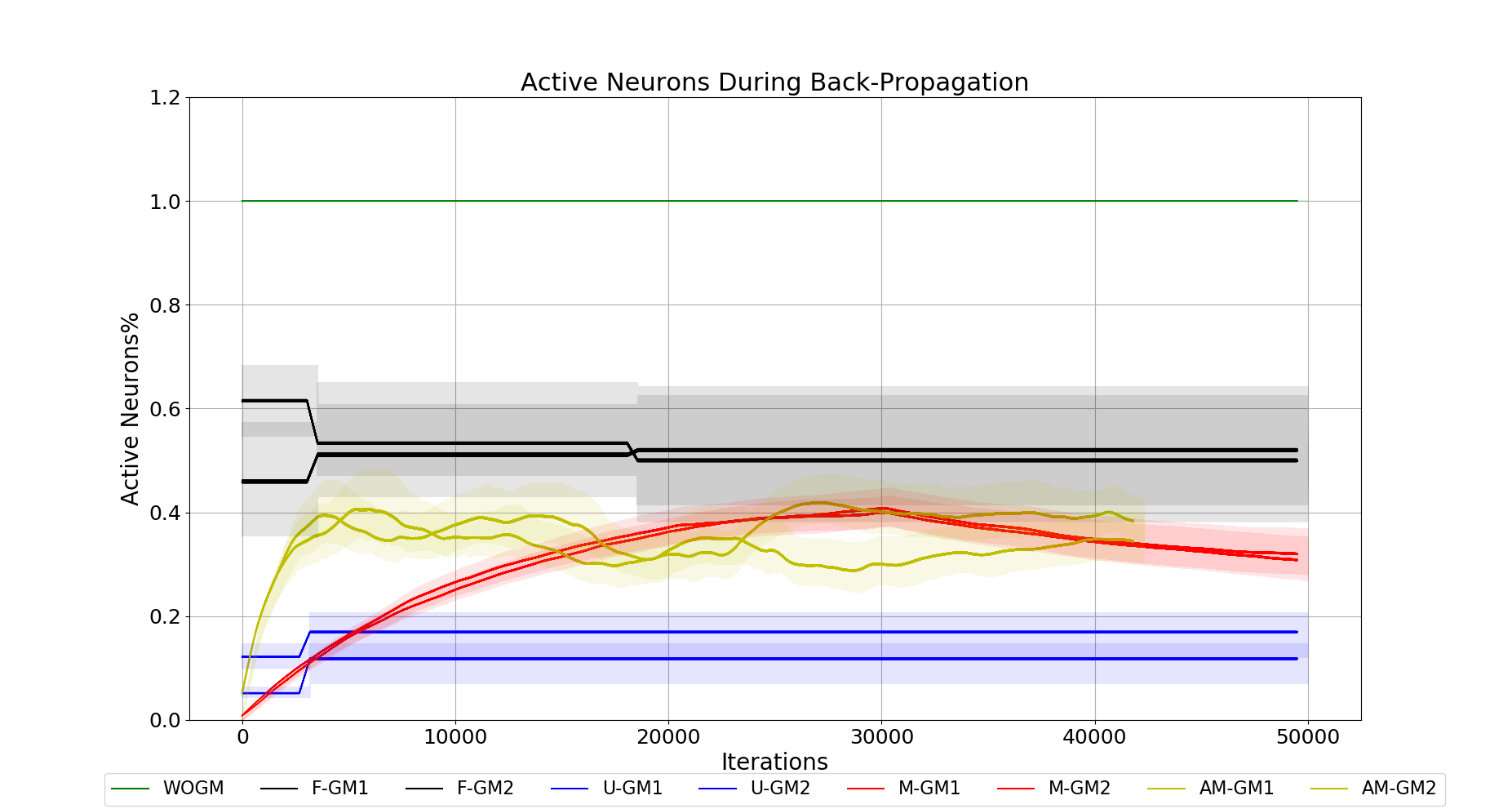}
    \caption{Active neurons in back-prop}
    \label{fig:act_neurons}
\end{figure}

\begin{figure}[t]
    \centering
    \includegraphics[width=0.5\textwidth, align=c]{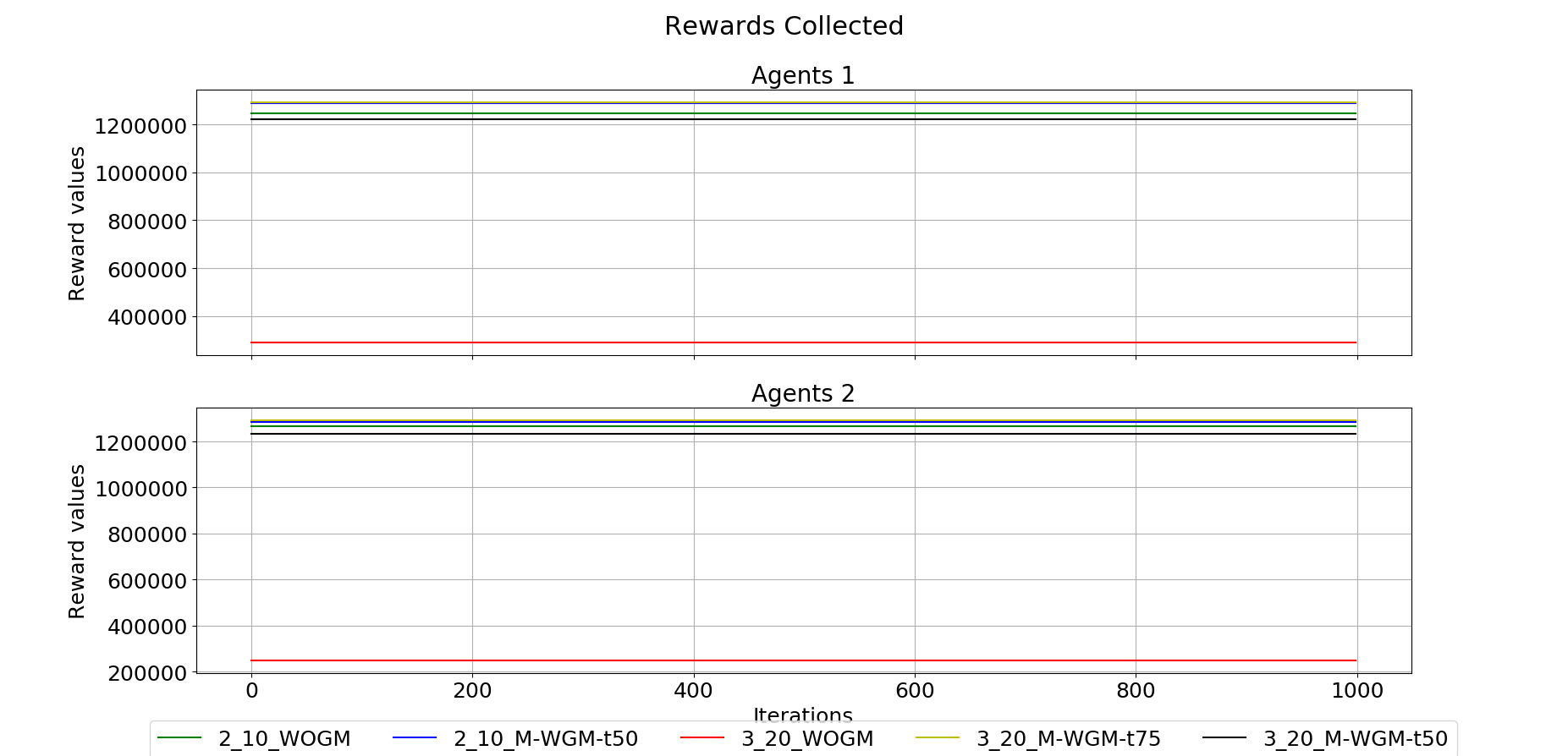}
    \caption{Rewards collected by the agents of different network sizes }
    \label{fig:nn_rew}
\end{figure}

\begin{figure}[t]
    \centering
    \includegraphics[width=0.5\textwidth, align=c]{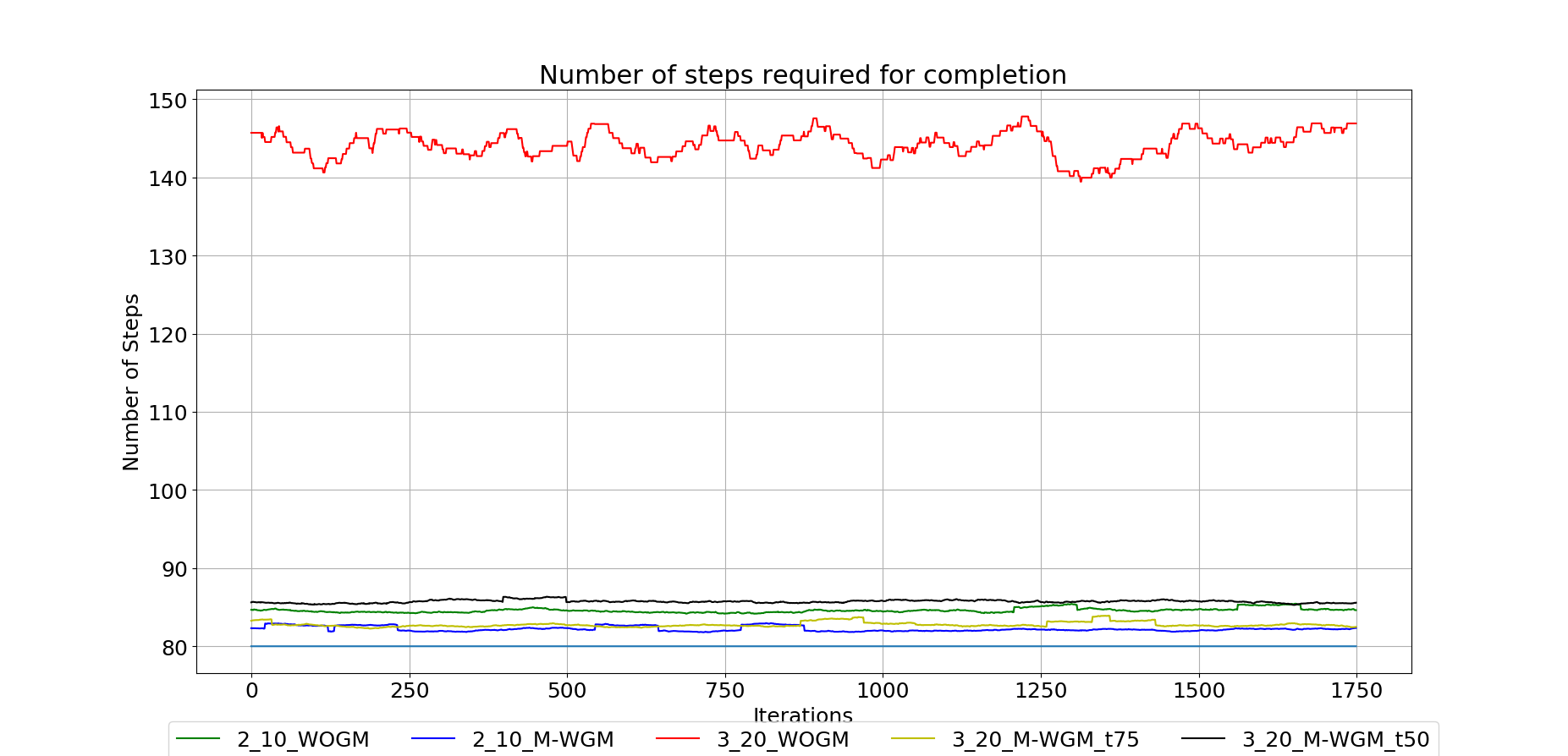}
    \caption{Output by different NN sizes}
    \label{fig:nn_output}
\end{figure}

\begin{figure}[t]
    \centering
    \includegraphics[scale=0.2, align=c]{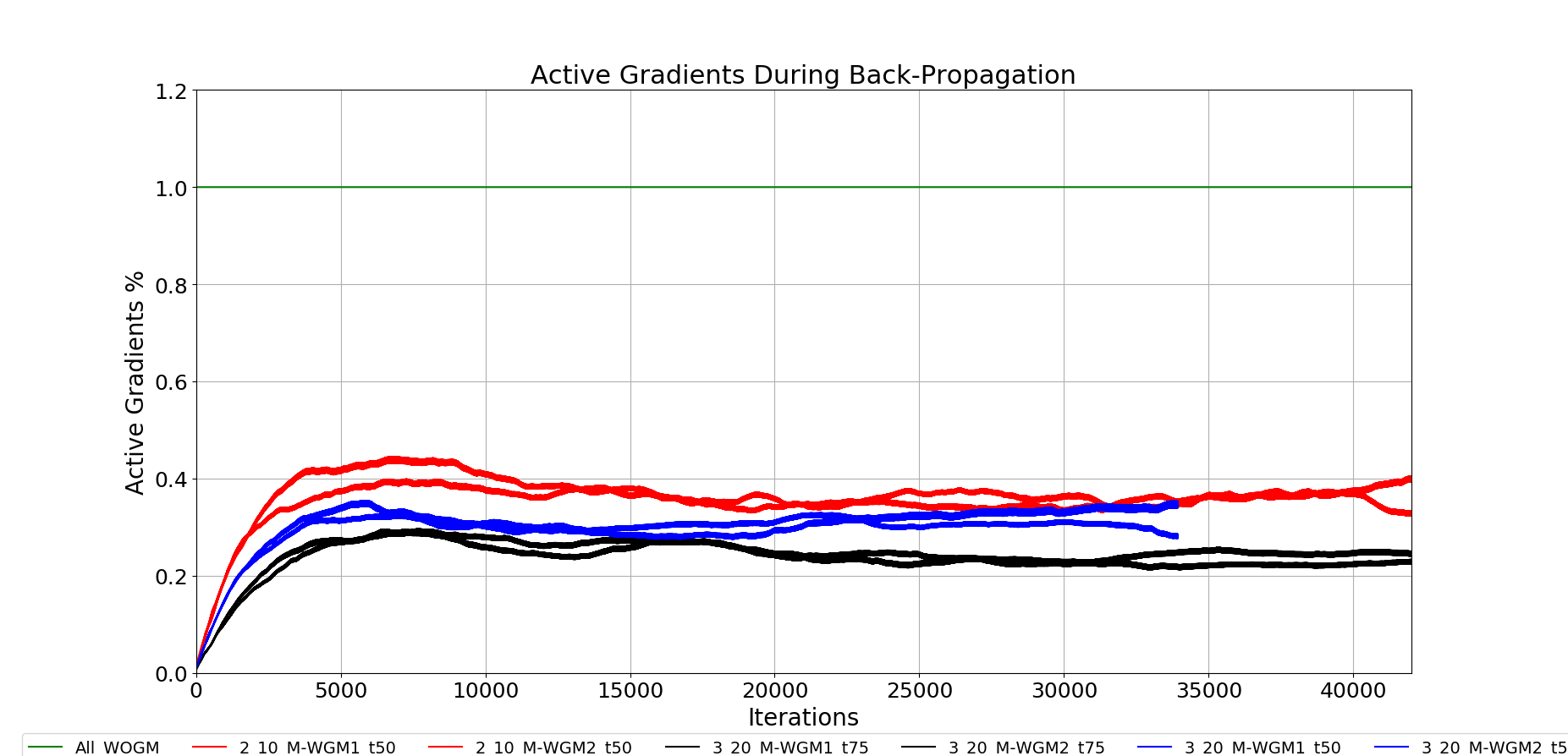}
    \caption{Comparison of active gradient percentage by network size}
    \label{fig:nn_act_perct}
\end{figure}

\textbf{Robustness to Choice of Network Size:}
Another important advantage of using the GM methods is the higher degree of freedom or robustness to the size of the network chosen. This is because the threshold-function ($\lambda\mu$) explained in Algorithm~\ref{alg:gm} adaptively selects only the required neurons for learning and ensures the learning is focused only on them. In  Figure.~\ref{fig:act_neurons}, the dynamic selection of the amount active neurons from all the GM methods are illustrated over the training progress. This dynamic selection accelerates the learning process while removing the need for hyper-parameter tuning for the number of neurons in the DNN. To provide additional evidence for this phenomenon, we trained the same multi-robot co-ordination problem with a randomly chosen bigger network size (3-layers, 20-neurons per layer) with the  M-WGM algorithm. Three simulations were made one without any GM method, one with M-WGM (threshold - 0.5) and one with M-WGM (threshold - 0.75).  As illustrated  in Figure~\ref{fig:nn_rew}, the rewards collected by the WOGM method with more parameters,  is considerably less to all the M-WGM methods when the  network size is changed. The drop in performance is substantially less in the GM algorithms. Furthermore, Figure~\ref{fig:nn_output} illustrates the drastic increase in  the number of steps required for the  WOGM method  to achieve the work-piece transportation goal. This shows the robustness to the change in the size of the network.  Fig.~\ref{fig:nn_act_perct} illustrates the automatic adjustment on the amount of active weights  in the M-WGM methods. It can be observed  that the for the same learning factor ($\lambda$)value  of 50\%, the quantum of gradients back-propagated in the smaller network is higher than in the bigger network, further proving the automatic usable network size adjustment capability of the algorithm.

\begin{figure}[t]
    \centering
    \includegraphics[scale=0.2, align=c]{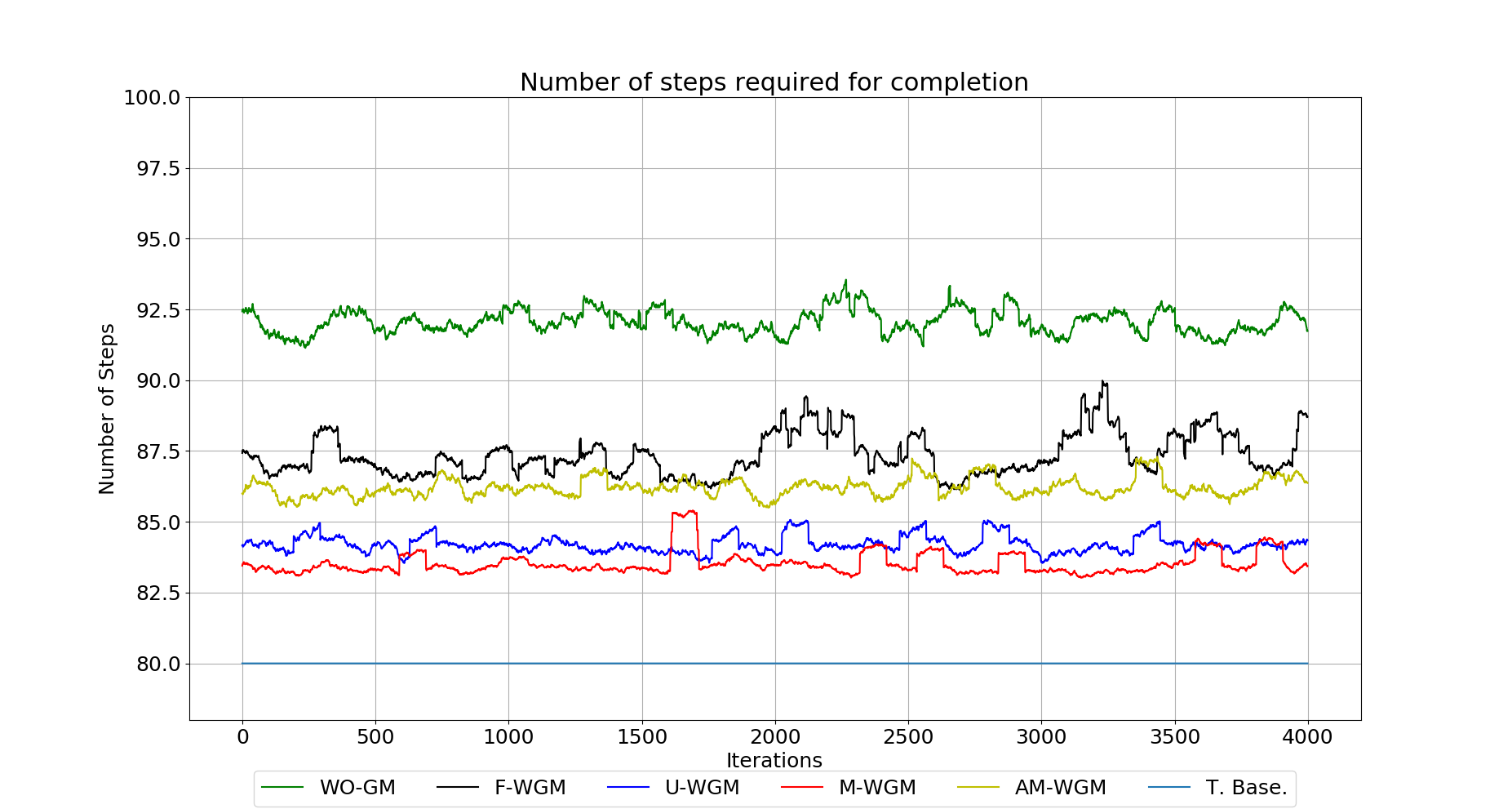}
    \caption{Steps taken to complete the given target}
    \label{fig:task_time}
\end{figure}

\begin{figure}[t]
    \centering
    \includegraphics[scale=0.2, align=c]{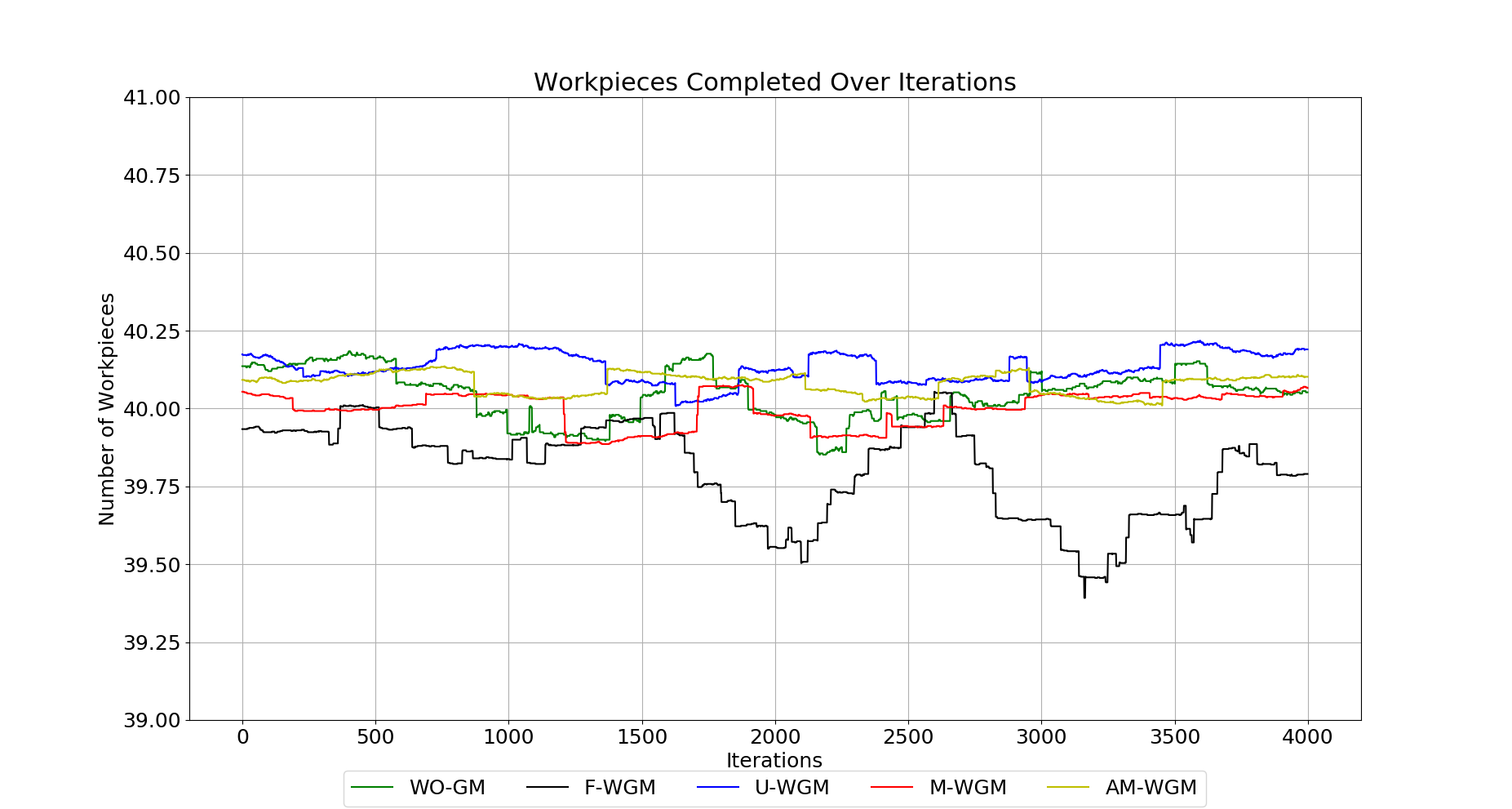}
    \caption{Work-pieces completed in each episode}
    \label{fig:wp_output}
\end{figure}

\textbf{Task Time:}
Task time is the amount of steps required by the agents to move the 20 work-pieces through the production system. The figures show the episode wise steps taken (Figure~\ref{fig:task_time}) and jobs completed (Figure~\ref{fig:wp_output}). WOGM performs the worst in terms of number of steps required and is also not stable in work completion. F-WGM improves the task time while still being a bit unstable in work-piece completion. U-WGM provides a very stable output at the same time reducing the task time further. While M-WGM provides the best task completion time and is stable enough for deployment, AM-WGM provides the combination of stability and task time completion. This also reflects in the amount of rewards collected.

\subsection{MuJoCo}
\subsubsection{Environment Description:}
The MuJoCo engine facilitates accurate and fast simulations of physical systems for research and development. This engine, wrapped inside the OpenAI Gym environment, provides for a frequently used~\cite{haarnoja2018soft,fujimoto2018addressing} training and testing benchmark environment in the domain of RL. The already established baseline performance in the form of PPO algorithm helps in the direct comparison of the effect of introducing GM methods. We test the M-WGM and AM-WGM on four randomly selected MuJoCo environments i.e. Half Cheetah, Ant, Humanoid and Inverted Double Pendulum (IDP). Each algorithm was run on the four environments with 10 random seed initialization. The learning setup is the same as in~\cite{Schulman.2017}, if not stated otherwise. The hyperparameter for all the different algorithms are shown in table~\ref{tab:hpy-Mujoco-ppo}.

\subsubsection{Results}
\begin{figure}[t]
    \centering
    \includegraphics[scale=0.2, align=c]{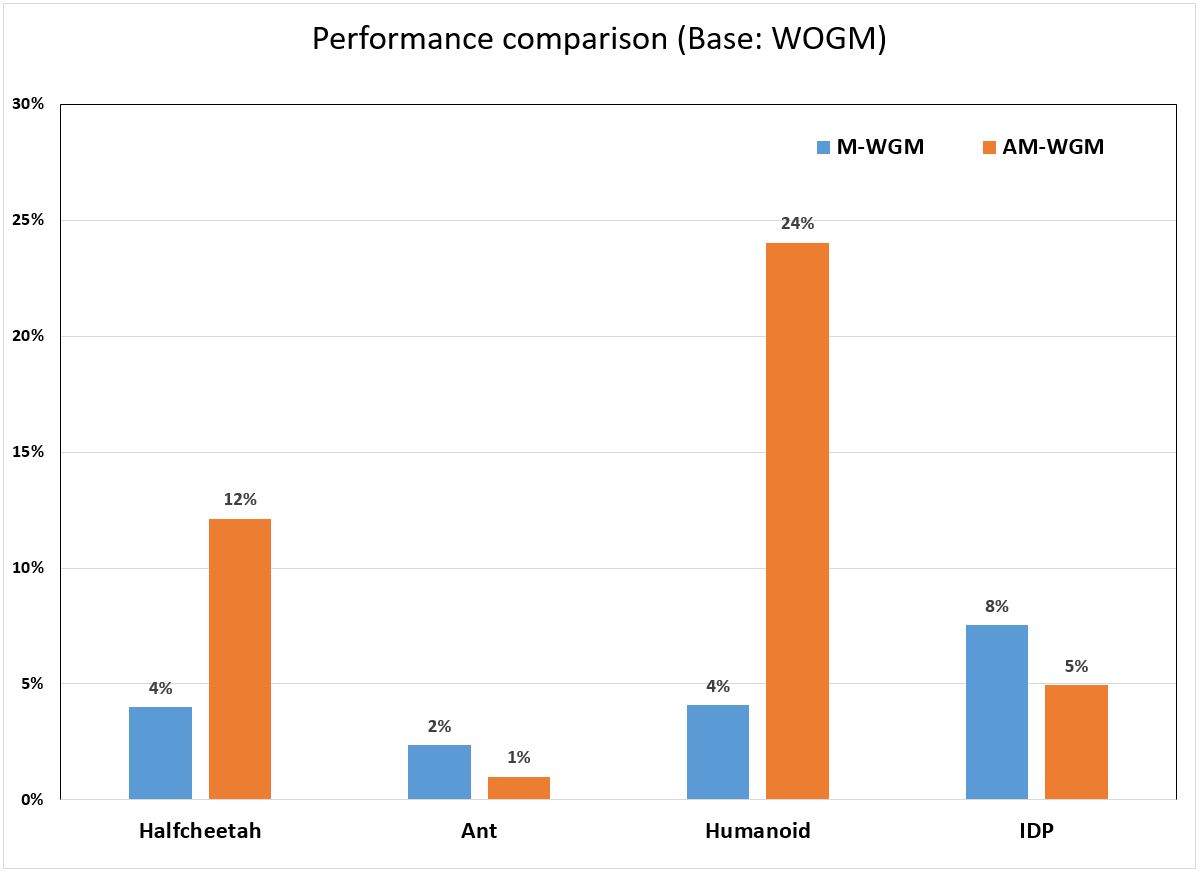}
    \caption{Percent Change in the performance of the M-WGM and AM-WGM with WOGM as baseline }
    \label{fig:mjc_rews}
\end{figure}

Since we are reducing the gradients back-propagated, the hyperparameters of the PPO algorithm are modified to reflect that and take advantage of the reduced variances. For example, the learning rate is increased compared to the original paper. This does not destabilize the learning process like in the vanilla PPO implementation due to the inherent variance reduction capabilities of the M-WGM and AM-WGM algorithms. It should be noted that we have also not used the global gradient clipping implemented in the vanilla PPO. The GM implementation provides a better layer-wise control over norm of the gradients. As illustrated in Fig.~\ref{fig:mjc_rews}, during our trials both the GM methods performed better than WOGM. The M-WGM and AM-WGM algorithms both performed on average better in all of the four games across the 10 random seeds. It should be noted that the AM-WGM provides the best normalized improvement over the WOGM. The final  scores  with the maximum average reward collected  are presented  in Table~\ref{tab:mujoco_scores}.

\begin{table}
\caption{Reward in MuJoCo environment}
\begin{center}
\begin{tabular}{ |c|c|c|c| }
\hline
Environment & PPO & M-WGM & AM-WGM\\
\hline
Half Cheetah&3600$\pm$1447&3744$\pm$1621&4037$\pm$1785\\
Ant&3151$\pm$584&3225$\pm$611&3183$\pm$758\\
Humanoid&720$\pm$381&750$\pm$658&893$\pm$1007\\
IDP&7583$\pm$1151&8154$\pm$1063&8364$\pm$959\\
\hline
\end{tabular}
\label{tab:mujoco_scores}
\end{center}
\end{table}

\begin{table}
\caption{Hyperparameter Values for PPO in MuJoCo}
\begin{center}
\begin{tabular}{ |c|c|c| } 
\hline
Algorithm & Hyperparameter & Value\\
\hline
WOGM & Learning rate & 2.5e-4\\
WOGM & Hidden Units & 64\\
M- \& AM-WGM & Learning rate & 3e-4\\
M- \& AM-WGM & Hidden Units & 96\\
WOGM & k-epoch updates & 4\\
M- \& AM-WGM & k-epoch updates & 5\\
M- \& AM-WGM & Momentum Value ($\zeta$) & 0.99, 0.9995\\
M- \& AM-WGM & Threshold ($\lambda$) & 0.5\\
M- \& AM-WGM & Global Gradient clipping & False\\
M- \& AM-WGM & Momentum Matrix($M_\zeta$) Init & 1\\
AM-WGM & Threshold Change ($\alpha_\zeta$) & 0.05\\
AM-WGM & Adaptive start from & 150\\
AM-WGM & Adaptive start for & 50\\
\hline
\end{tabular}
\label{tab:hpy-Mujoco-ppo}
\end{center}
\end{table}

\subsection{Atari}
\subsubsection{Environment Description}
The Atari games were first introduced in~\cite{Bellemare.2013} to aid the development of general domain independent AI technology. The environment also provides for a baseline where previous algorithms have been tested. We test a total 10 games, 6 of which were randomly selected (Battlezone, Frostbite, Gopher, Kangaroo, Timepilot, and Zaxxon). The other 4 (Montezuma's Revenge, Pitfall, Skiing, and Solaris) are selected to specifically to test the long term credit assignment problem of the algorithms. We use the ram information as input to the network with no frames being skipped. The 10 games were run on the three algorithms (WOGM, M-WGM, and AM-WGM) over 5 random seed initialization. The learning setup is the same as in~\cite{Schulman.2017}, if not stated otherwise. The hyperparameter for all the different algorithms are shown in table~\ref{tab:hpy-Atari-ppo}.

\subsubsection{Results}

\begin{table}
\caption{Hyperparameter Values for PPO in Atari games}
\begin{center}
\begin{tabular}{ |c|c|c| } 
\hline
Algorithm & Hyperparameter & Value\\
\hline
WOGM & Learning rate & 2.5e-4\\
WOGM & Hidden Units & 64\\
M- \& AM-WGM & Learning rate & 4e-4\\
M- \& AM-WGM & Hidden Units & 96\\
M- \& AM-WGM & Momentum Value ($\zeta$) & 0.999\\
M- \& AM-WGM & Threshold ($\lambda$) & 0.5\\
M- \& AM-WGM & Global Gradient clipping & False\\
M- \& AM-WGM & Momentum Matrix($M_\zeta$) Init & 0\\
AM-WGM & Threshold Change ($\alpha_\zeta$) & 0.1\\
AM-WGM & Adaptive start from & 2000\\
AM-WGM & Adaptive start for & 1000\\
\hline
\end{tabular}
\label{tab:hpy-Atari-ppo}
\end{center}
\end{table}

\begin{figure}[t]
    \centering
    \includegraphics[scale=0.2, align=c]{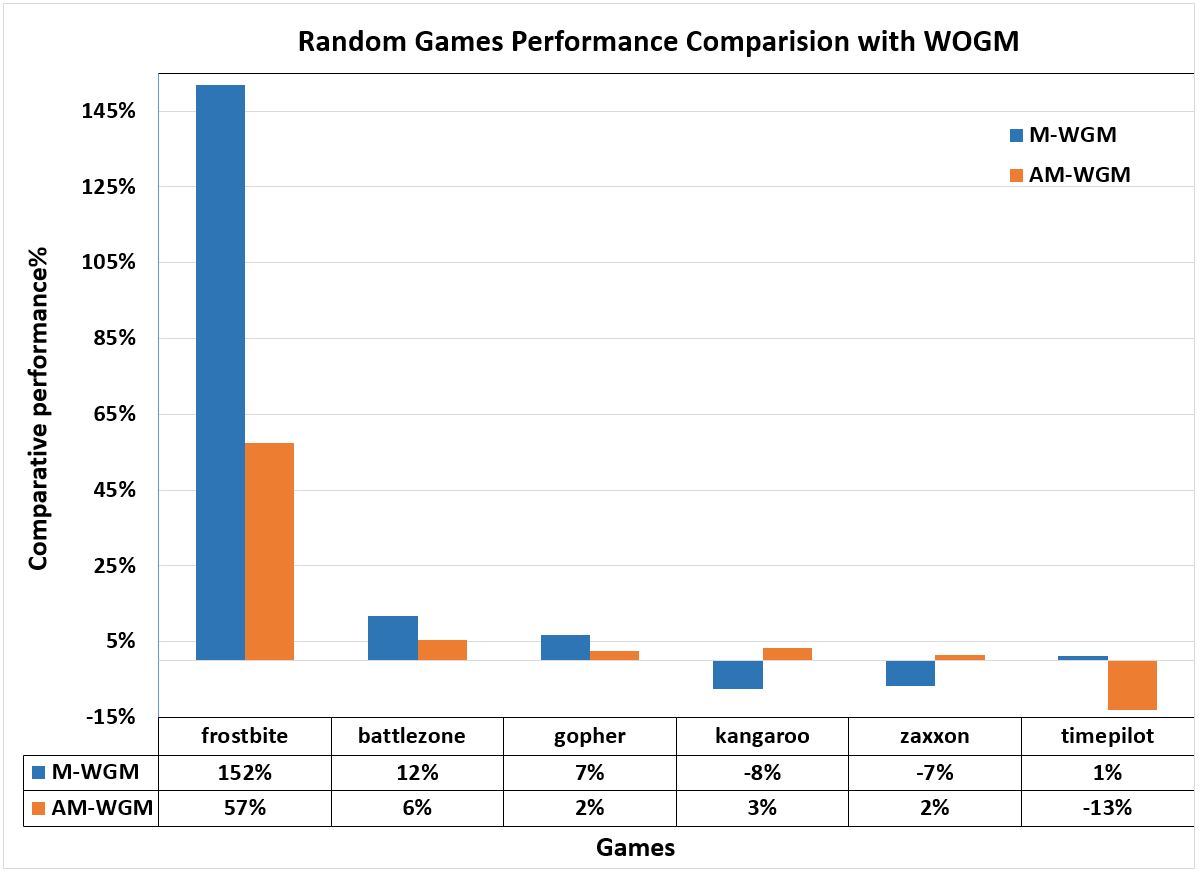}
    \caption{Percentage Performance improvement of the proposed methods in 6 randomly selected games}
    \label{fig:atari_rews_1}
\end{figure}

\begin{figure}[t]
    \centering
    \includegraphics[scale=0.2, align=c]{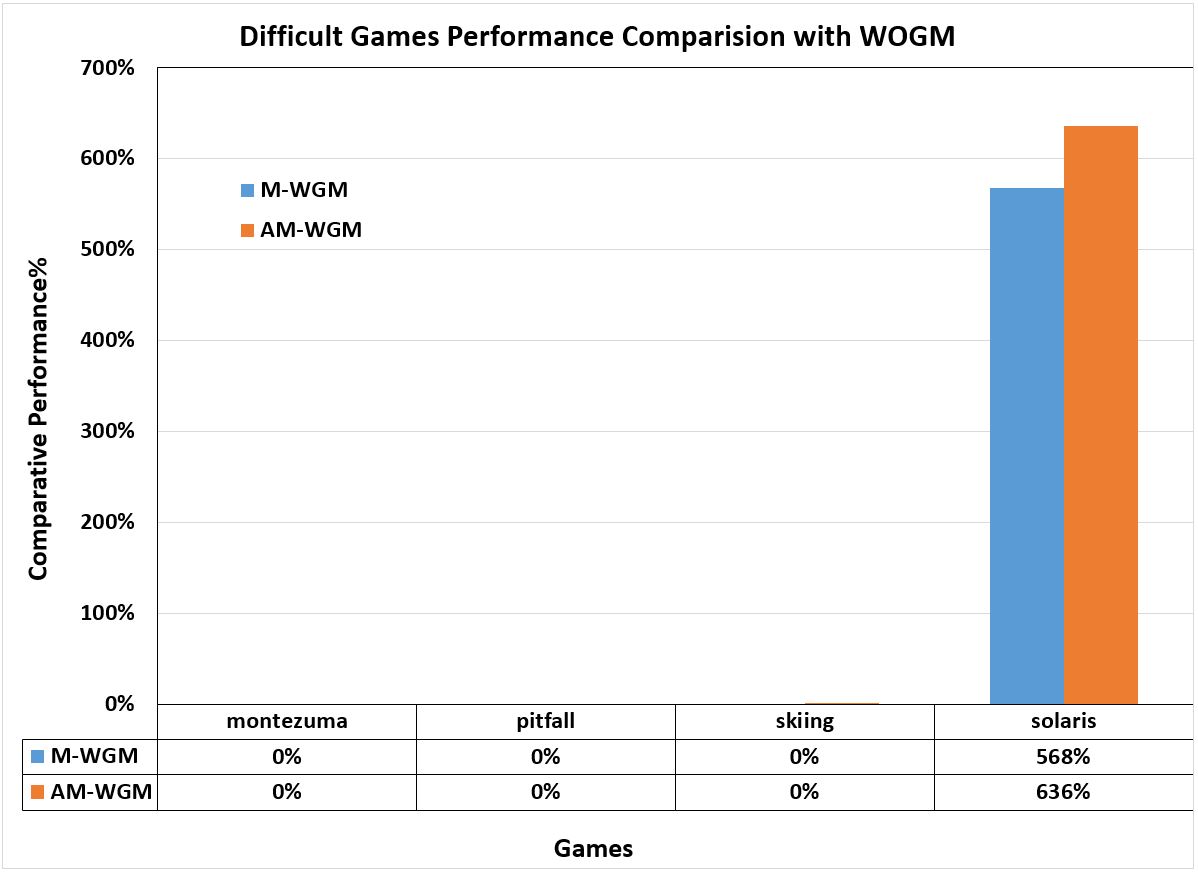}
    \caption{Performance improvement in difficult games}
    \label{fig:atari_rews_2}
\end{figure}

As with the implementation in MuJoCo environment, we use a higher learning rate and do not use the global clipping used in the vanilla PPO. We also found that increasing the k-epoch update in AM-WGM increases its performance significantly. As shown in the Figure~\ref{fig:atari_rews_1}, the M-WGM method performs better than WOGM in 4 out of the 6 random games while AM-WGM performs better in 5 out of the 6 random games. There was no performance improvement for the algorithms in the difficult games except in Solaris, where there is a drastic improvement made by the GM algorithms as shown in Figure~\ref{fig:atari_rews_2}.

\section{Conclusion}\label{sec:conclusion}
We propose four novel neural network training methodologies called Gradient Monitoring in Reinforcement Learning,  for a more  robust and faster training progress. The proposed methods incorporate a targeted training procedure in neural network by systematically reducing the gradient variance and additional trust-region constraint for the policy updates. The adaptive momentum method helps the network to chose the optimal number of parameters required for a particular training step based on the feedback from the rewards collected. This results in the training algorithm being robust to the selection of the size of the network. The proposed methods on average, outperform the standard A2C in the multi-robot co-operation application and the standard   PPO algorithm in the Mujoco and Atari environment.

A potential limitation of  the F-WGM and UF-WGM methods is the occurrence of peaks in the the gradient during training  which can sometimes disturb the learning process. Another limitation of the AM-WGM is the selection of the hyperparameter $\eta_{start}$ .This can be eliminated by  using feedback from the reward collection during training. This will part of the future work. Subsequent research will also focus on the performance improvement of the RL agent to generalize to unseen environment setup like in CoinRun~\cite{Cobbe.2019}, application to model free on-policy RL algorithms like trust region policy optimization~\cite{Schulman.2015} and model free off-policy RL algorithms like deep deterministic policy gradient \cite{TimothyP.Lillicrap.2016}.


\printbibliography

@inproceedings{zhu2017target,
  title={Target-driven visual navigation in indoor scenes using deep reinforcement learning},
  author={Zhu, Yuke and Mottaghi, Roozbeh and Kolve, Eric and Lim, Joseph J and Gupta, Abhinav and Fei-Fei, Li and Farhadi, Ali},
  booktitle={2017 IEEE international conference on robotics and automation (ICRA)},
  pages={3357--3364},
  year={2017},
  organization={IEEE}
}

@inproceedings{haarnoja2018soft,
  title={Soft Actor-Critic: Off-Policy Maximum Entropy Deep Reinforcement Learning with a Stochastic Actor},
  author={Haarnoja, Tuomas and Zhou, Aurick and Abbeel, Pieter and Levine, Sergey},
  booktitle={International Conference on Machine Learning},
  pages={1861--1870},
  year={2018}
}

@inproceedings{fujimoto2018addressing,
  title={Addressing Function Approximation Error in Actor-Critic Methods},
  author={Fujimoto, Scott and Hoof, Herke and Meger, David},
  booktitle={International Conference on Machine Learning},
  pages={1587--1596},
  year={2018}
}

@inproceedings{Anschel.2017,
 author = {Anschel, Oron and Baram, Nir and Shimkin, Nahum},
 title = {Averaged-DQN: Variance Reduction and Stabilization for Deep Reinforcement Learning},
 pages = {176--185},
 publisher = {JMLR.org},
 series = {ICML'17},
 booktitle = {Proceedings of the 34th International Conference on Machine Learning - Volume 70},
 year = {2017}
}

@article{Applegate.1991,
 author = {Applegate, David and Cook, William},
 year = {1991},
 title = {A Computational Study of the Job-Shop Scheduling Problem},
 pages = {149--156},
 volume = {3},
 number = {2},
 issn = {0899-1499},
 journal = {ORSA Journal on Computing}
}

@inproceedings{Arora.2018b,
 author = {Arora, Sanjeev and Cohen, Nadav and Hazan, Elad},
 title = {On the Optimization of Deep Networks: Implicit Acceleration by Overparameterization},
 pages = {244--253},
 volume = {80},
 publisher = {PMLR},
 series = {Proceedings of Machine Learning Research},
 booktitle = {Proceedings of the 35th International Conference on Machine Learning},
 year = {2018}
}

@inproceedings{Baker.2017,
 author = {Baker, Bowen and Gupta, Otkrist and Naik, Nikhil and Raskar, Ramesh},
 title = {Designing neural network architectures using reinforcement learning},
 booktitle = {International Conference on Learning Representations},
 year = {2017}
}

@inproceedings{BehnamNeyshabur.2019,
 author = {{Behnam Neyshabur} and {Zhiyuan Li} and {Srinadh Bhojanapalli} and {Yann LeCun} and {Nathan Srebro}},
 title = {The role of over-parametrization in generalization of neural networks},
 booktitle = {International Conference on Learning Representations},
 year = {2019}
}

@article{Belkin.2019,
 author = {Belkin, Mikhail and Hsu, Daniel and Ma, Siyuan and Mandal, Soumik},
 year = {2019},
 title = {Reconciling modern machine-learning practice and the classical bias-variance trade-off},
 pages = {15849--15854},
 volume = {116},
 number = {32},
 issn = {0027-8424},
 journal = {Proceedings of the National Academy of Sciences}
}

@article{Bellemare.2013,
 author = {Bellemare, Marc G. and Naddaf, Yavar and Veness, Joel and Bowling, Michael},
 year = {2013},
 title = {The arcade learning environment: An evaluation platform for general agents},
 pages = {253--279},
 volume = {47},
 journal = {Journal of Artificial Intelligence Research}
}

@article{Bergstra.2012,
 author = {Bergstra, James and Bengio, Yoshua},
 year = {2012},
 title = {Random search for hyper-parameter optimization},
 pages = {281--305},
 volume = {13},
 number = {Feb},
 journal = {Journal of Machine Learning Research}
}

@inproceedings{Bergstra.2013,
 author = {Bergstra, J. and Yamins, D. and Cox, D. D.},
 title = {Making a Science of Model Search: Hyperparameter Optimization in Hundreds of Dimensions for Vision Architectures},
 pages = {I--115--I--123},
 publisher = {JMLR.org},
 series = {ICML'13},
 booktitle = {Proceedings of the 30th International Conference on International Conference on Machine Learning - Volume 28},
 year = {2013}
}

@incollection{Blalock.2020,
 author = {Blalock, Davis and {Gonzalez Ortiz}, Jose Javier and Frankle, Jonathan and Guttag, John},
 title = {What is the State of Neural Network Pruning?},
 pages = {129--146},
 booktitle = {Proceedings of Machine Learning and Systems 2020},
 year = {2020}
}

@inproceedings{Brutzkus.2019,
 author = {Brutzkus, Alon and Globerson, Amir},
 title = {Why do Larger Models Generalize Better? A Theoretical Perspective via the XOR Problem},
 pages = {822--830},
 booktitle = {International Conference on Machine Learning},
 year = {2019}
}

@inproceedings{Chadha.2019c,
 author = {Chadha, Gavneet Singh and Meydani, Elnaz and Schwung, Andreas},
 title = {Regularizing Neural Networks with Gradient Monitoring},
 pages = {196--205},
 booktitle = {INNS Big Data and Deep Learning conference},
 year = {2019}
}

@inproceedings{Cobbe.2019,
 author = {Cobbe, Karl and Klimov, Oleg and Hesse, Chris and Kim, Taehoon and Schulman, John},
 title = {Quantifying Generalization in Reinforcement Learning},
 pages = {1282--1289},
 booktitle = {International Conference on Machine Learning},
 year = {2019}
}

@inproceedings{Conti.2018,
 author = {Conti, Edoardo and Madhavan, Vashisht and Such, Felipe Petroski and Lehman, Joel and Stanley, Kenneth and Clune, Jeff},
 title = {Improving exploration in evolution strategies for deep reinforcement learning via a population of novelty-seeking agents},
 pages = {5027--5038},
 booktitle = {Advances in neural information processing systems},
 year = {2018}
}

@inproceedings{DiederikP.Kingma.2015,
 author = {{Diederik P. Kingma} and {Jimmy Ba}},
 title = {Adam: A Method for Stochastic Optimization},
 booktitle = {3rd International Conference on Learning Representations, ICLR 2015},
 year = {2015}
}

@inproceedings{E.Todorov.2012,
 author = {{E. Todorov} and {T. Erez} and {Y. Tassa}},
 title = {MuJoCo: A physics engine for model-based control},
 pages = {5026--5033},
 booktitle = {2012 IEEE/RSJ International Conference on Intelligent Robots and Systems},
 year = {2012}
}

@inproceedings{Glorot.2010,
 author = {Glorot, Xavier and Bengio, Yoshua},
 title = {Understanding the difficulty of training deep feedforward neural networks},
 pages = {249--256},
 booktitle = {Proceedings of the thirteenth international conference on artificial intelligence and statistics},
 year = {2010}
}

@inproceedings{Glorot.2011,
 author = {Glorot, Xavier and Bordes, Antoine and Bengio, Yoshua},
 title = {Deep Sparse Rectifier Neural Networks},
 pages = {275},
 volume = {15},
 booktitle = {Aistats},
 year = {2011}
}

@article{Gomez.2019,
 author = {Gomez, Aidan N. and Zhang, Ivan and Swersky, Kevin and Gal, Yarin and Hinton, Geoffrey E.},
 year = {2019},
 title = {Learning Sparse Networks Using Targeted Dropout},
 journal = {arXiv preprint arXiv:1905.13678}
}

@book{Goodfellow.2016,
 author = {Goodfellow, Ian and Bengio, Yoshua and Courville, Aaron},
 year = {2016},
 title = {Deep Learning},
 publisher = {{MIT Press}}
}

@article{Gori.1992,
 author = {Gori, Marco and Tesi, Alberto},
 year = {1992},
 title = {On the problem of local minima in backpropagation},
 pages = {76--86},
 number = {1},
 journal = {IEEE Transactions on Pattern Analysis {\&} Machine Intelligence}
}

@article{Greensmith.2004,
 author = {Greensmith, Evan and Bartlett, Peter L. and Baxter, Jonathan},
 year = {2004},
 title = {Variance reduction techniques for gradient estimates in reinforcement learning},
 pages = {1471--1530},
 volume = {5},
 number = {Nov},
 journal = {Journal of Machine Learning Research}
}

@inproceedings{Gu.2017,
 author = {Gu, Shixiang and Holly, Ethan and Lillicrap, Timothy and Levine, Sergey},
 title = {Deep reinforcement learning for robotic manipulation with asynchronous off-policy updates},
 pages = {3389--3396},
 booktitle = {2017 IEEE international conference on robotics and automation (ICRA)},
 year = {2017}
}

@inproceedings{Han.2015,
 author = {Han, Song and Pool, Jeff and Tran, John and Dally, William},
 title = {Learning both weights and connections for efficient neural network},
 pages = {1135--1143},
 booktitle = {Advances in neural information processing systems},
 year = {2015}
}

@inproceedings{Hassibi.1993,
 author = {Hassibi, Babak and Stork, David G.},
 title = {Second order derivatives for network pruning: Optimal brain surgeon},
 pages = {164--171},
 booktitle = {Advances in neural information processing systems},
 year = {1993}
}

@inproceedings{He.2015,
 author = {He, Kaiming and Zhang, Xiangyu and Ren, Shaoqing and Sun, Jian},
 title = {Delving deep into rectifiers: Surpassing human-level performance on imagenet classification},
 pages = {1026--1034},
 booktitle = {Proceedings of the IEEE international conference on computer vision},
 year = {2015}
}

@article{Hochreiter.1998,
 author = {Hochreiter, Sepp},
 year = {1998},
 title = {The vanishing gradient problem during learning recurrent neural nets and problem solutions},
 pages = {107--116},
 volume = {6},
 number = {02},
 journal = {International Journal of Uncertainty, Fuzziness and Knowledge-Based Systems}
}

@inproceedings{HongziMao.2019,
 author = {{Hongzi Mao} and {Shaileshh Bojja Venkatakrishnan} and {Malte Schwarzkopf} and {Mohammad Alizadeh}},
 title = {Variance Reduction for Reinforcement Learning in Input-Driven Environments},
 booktitle = {International Conference on Learning Representations},
 year = {2019}
}

@inproceedings{Igl.2019,
 author = {Igl, Maximilian and Ciosek, Kamil and Li, Yingzhen and Tschiatschek, Sebastian and Zhang, Cheng and Devlin, Sam and Hofmann, Katja},
 title = {Generalization in reinforcement learning with selective noise injection and information bottleneck},
 pages = {13956--13968},
 booktitle = {Advances in Neural Information Processing Systems},
 year = {2019}
}

@inproceedings{Ioffe.2015b,
 author = {Ioffe, Sergey and Szegedy, Christian},
 title = {Batch Normalization: Accelerating Deep Network Training by Reducing Internal Covariate Shift},
 pages = {448--456},
 publisher = {JMLR.org},
 series = {ICML'15},
 booktitle = {Proceedings of the 32nd International Conference on International Conference on Machine Learning - Volume 37},
 year = {2015}
}

@article{Jaderberg.2017,
 author = {Jaderberg, Max and Dalibard, Valentin and Osindero, Simon and Czarnecki, Wojciech M. and Donahue, Jeff and Razavi, Ali and Vinyals, Oriol and Green, Tim and Dunning, Iain and Simonyan, Karen and others},
 year = {2017},
 title = {Population based training of neural networks},
 journal = {arXiv preprint arXiv:1711.09846}
}

@inproceedings{JingzhaoZhang.2020,
 author = {{Jingzhao Zhang} and {Tianxing He} and {Suvrit Sra} and {Ali Jadbabaie}},
 title = {Why Gradient Clipping Accelerates Training: A Theoretical Justification for Adaptivity},
 booktitle = {International Conference on Learning Representations},
 year = {2020}
}

@article{Justesen.2018,
 author = {Justesen, Niels and Torrado, Ruben Rodriguez and Bontrager, Philip and Khalifa, Ahmed and Togelius, Julian and Risi, Sebastian},
 year = {2018},
 title = {Illuminating generalization in deep reinforcement learning through procedural level generation},
 journal = {arXiv preprint arXiv:1806.10729}
}

@inproceedings{Kingma.2015b,
 author = {Kingma, Durk P. and Salimans, Tim and Welling, Max},
 title = {Variational dropout and the local reparameterization trick},
 pages = {2575--2583},
 booktitle = {Advances in neural information processing systems},
 year = {2015}
}

@inproceedings{Lample.2017,
 author = {Lample, Guillaume and Chaplot, Devendra Singh},
 title = {Playing FPS Games with Deep Reinforcement Learning},
 pages = {2140--2146},
 publisher = {{AAAI Press}},
 series = {AAAI'17},
 booktitle = {Proceedings of the Thirty-First AAAI Conference on Artificial Intelligence},
 year = {2017}
}

@inproceedings{Lawrence.1997,
 author = {Lawrence, Steve and Giles, C. Lee and Tsoi, Ah Chung},
 title = {Lessons in neural network training: Overfitting may be harder than expected},
 pages = {540--545},
 booktitle = {AAAI/IAAI},
 year = {1997}
}

@incollection{LeCun.1990b,
 author = {{Le Cun}, Yann and Denker, John S. and Solla, Sara A.},
 title = {Optimal Brain Damage},
 pages = {598--605},
 publisher = {{Morgan Kaufmann Publishers Inc}},
 booktitle = {Advances in Neural Information Processing Systems 2},
 year = {1990},
 address = {San Francisco, CA, USA}
}

@article{Li.2017b,
 author = {Li, Lisha and Jamieson, Kevin and DeSalvo, Giulia and Rostamizadeh, Afshin and Talwalkar, Ameet},
 year = {2017},
 title = {Hyperband: A Novel Bandit-Based Approach to Hyperparameter Optimization},
 pages = {6765--6816},
 volume = {18},
 number = {1},
 issn = {1532-4435},
 journal = {J. Mach. Learn. Res.}
}

@article{Mnih.2015,
 author = {Mnih, Volodymyr and Kavukcuoglu, Koray and Silver, David and Rusu, Andrei A. and Veness, Joel and Bellemare, Marc G. and Graves, Alex and Riedmiller, Martin and Fidjeland, Andreas K. and Ostrovski, Georg and Petersen, Stig and Beattie, Charles and Sadik, Amir and Antonoglou, Ioannis and King, Helen and Kumaran, Dharshan and Wierstra, Daan and Legg, Shane and Hassabis, Demis},
 year = {2015},
 title = {Human-level control through deep reinforcement learning},
 pages = {529--533},
 volume = {518},
 number = {7540},
 journal = {Nature}
}

@inproceedings{Mnih.2016,
 author = {Mnih, Volodymyr and Badia, Adria Puigdomenech and Mirza, Mehdi and Graves, Alex and Lillicrap, Timothy and Harley, Tim and Silver, David and Kavukcuoglu, Koray},
 title = {Asynchronous methods for deep reinforcement learning},
 pages = {1928--1937},
 booktitle = {International conference on machine learning},
 year = {2016}
}

@inproceedings{Mohamed.2015,
 author = {Mohamed, Shakir and Rezende, Danilo Jimenez},
 title = {Variational information maximisation for intrinsically motivated reinforcement learning},
 pages = {2125--2133},
 booktitle = {Advances in neural information processing systems},
 year = {2015}
}

@inproceedings{NamhoonLee.2019,
 author = {{Namhoon Lee} and {Thalaiyasingam Ajanthan} and {Philip Torr}},
 title = {SNIP: SINGLE-SHOT NETWORK PRUNING BASED ON CONNECTION SENSITIVITY},
 booktitle = {International Conference on Learning Representations},
 year = {2019}
}

@inproceedings{NamhoonLee.2020,
 author = {{Namhoon Lee} and {Thalaiyasingam Ajanthan} and {Stephen Gould} and {Philip H. S. Torr}},
 title = {A Signal Propagation Perspective for Pruning Neural Networks at Initialization},
 booktitle = {International Conference on Learning Representations},
 year = {2020}
}

@inproceedings{Pascanu.2013,
 author = {Pascanu, Razvan and Mikolov, Tomas and Bengio, Yoshua},
 title = {On the difficulty of training recurrent neural networks},
 pages = {1310--1318},
 booktitle = {2013 International conference on machine learning},
 year = {2013}
}

@inproceedings{Paszke.2019b,
 author = {Paszke, Adam and Gross, Sam and Massa, Francisco and Lerer, Adam and Bradbury, James and Chanan, Gregory and Killeen, Trevor and Lin, Zeming and Gimelshein, Natalia and Antiga, Luca and others},
 title = {PyTorch: An imperative style, high-performance deep learning library},
 pages = {8024--8035},
 booktitle = {Advances in Neural Information Processing Systems},
 year = {2019}
}

@inproceedings{Pathak.2017,
 author = {Pathak, Deepak and Agrawal, Pulkit and Efros, Alexei A. and Darrell, Trevor},
 title = {Curiosity-Driven Exploration by Self-Supervised Prediction},
 pages = {2778--2787},
 publisher = {JMLR.org},
 series = {ICML'17},
 booktitle = {Proceedings of the 34th International Conference on Machine Learning - Volume 70},
 year = {2017}
}

@article{Schaul.2019,
 author = {Schaul, Tom and Borsa, Diana and Ding, David and Szepesvari, David and Ostrovski, Georg and Dabney, Will and Osindero, Simon},
 year = {2019},
 title = {Adapting Behaviour for Learning Progress},
 journal = {arXiv preprint arXiv:1912.06910}
}

@inproceedings{Schulman.2015,
 author = {Schulman, John and Levine, Sergey and Abbeel, Pieter and Jordan, Michael and Moritz, Philipp},
 title = {Trust region policy optimization},
 pages = {1889--1897},
 booktitle = {International conference on machine learning},
 year = {2015}
}

@article{Schulman.2017,
 author = {Schulman, John and Wolski, Filip and Dhariwal, Prafulla and Radford, Alec and Klimov, Oleg},
 year = {2017},
 title = {Proximal policy optimization algorithms},
 journal = {arXiv preprint arXiv:1707.06347}
}

@article{Silver.2017,
 author = {Silver, David and Schrittwieser, Julian and Simonyan, Karen and Antonoglou, Ioannis and Huang, Aja and Guez, Arthur and Hubert, Thomas and Baker, Lucas and Lai, Matthew and Bolton, Adrian and Chen, Yutian and Lillicrap, Timothy and Hui, Fan and Sifre, Laurent and {van den Driessche}, George and Graepel, Thore and Hassabis, Demis},
 year = {2017},
 title = {Mastering the game of Go without human knowledge},
 pages = {354--359},
 volume = {550},
 number = {7676},
 journal = {Nature}
}

@article{Srivastava.2014,
 author = {Srivastava, Nitish and Hinton, Geoffrey E. and Krizhevsky, Alex and Sutskever, Ilya and Salakhutdinov, Ruslan},
 year = {2014},
 title = {Dropout: a simple way to prevent neural networks from overfitting},
 pages = {1929--1958},
 volume = {15},
 number = {1},
 journal = {Journal of Machine Learning Research}
}

@inproceedings{T.Inoue.2017,
 author = {{T. Inoue} and {G. De Magistris} and {A. Munawar} and {T. Yokoya} and {R. Tachibana}},
 title = {Deep reinforcement learning for high precision assembly tasks},
 pages = {819--825},
 booktitle = {2017 IEEE/RSJ International Conference on Intelligent Robots and Systems (IROS)},
 year = {2017}
}

@incollection{TimothyP.Lillicrap.2016,
 author = {{Timothy P. Lillicrap} and {Jonathan J. Hunt} and {Alexander Pritzel} and {Nicolas Heess} and {Tom Erez} and {Yuval Tassa} and {David Silver} and {Daan Wierstra}},
 title = {Continuous control with deep reinforcement learning},
 booktitle = {2016 -- 4th International Conference on Learning}
}

@inproceedings{Wan.2013,
 author = {Wan, Li and Zeiler, Matthew and Zhang, Sixin and {Le Cun}, Yann and Fergus, Rob},
 title = {Regularization of neural networks using dropconnect},
 pages = {1058--1066},
 booktitle = {International conference on machine learning},
 year = {2013}
}

@inproceedings{XingyouSong.2020,
 author = {{Xingyou Song} and {Yiding Jiang} and {Stephen Tu} and {Yilun Du} and {Behnam Neyshabur}},
 title = {Observational Overfitting in Reinforcement Learning},
 booktitle = {International Conference on Learning Representations},
 year = {2020}
}

@inproceedings{Young.2015,
 author = {Young, Steven R. and Rose, Derek C. and Karnowski, Thomas P. and Lim, Seung-Hwan and Patton, Robert M.},
 title = {Optimizing Deep Learning Hyper-Parameters through an Evolutionary Algorithm},
 publisher = {{Association for Computing Machinery}},
 isbn = {9781450340069},
 series = {MLHPC '15},
 booktitle = {Proceedings of the Workshop on Machine Learning in High-Performance Computing Environments},
 year = {2015},
 address = {New York, NY, USA}
}

@book{Sutton.1998,
 author = {Sutton, Richard S. and Barto, Andrew G.},
 year = {1998},
 title = {Reinforcement learning: An introduction /   Richard S. Sutton and Andrew G. Barto},
 address = {Cambridge, Mass. and London},
 publisher = {{MIT Press}},
 isbn = {0262193981},
 series = {Adaptive computation and machine learning}
}

@article{Panait.2005,
 author = {Panait, Liviu and Luke, Sean},
 year = {2005},
 title = {Cooperative multi-agent learning: The state of the art},
 pages = {387--434},
 volume = {11},
 number = {3},
 journal = {Autonomous agents and multi-agent systems}
}

@article{Caruana.1997,
 author = {Caruana, Rich},
 year = {1997},
 title = {Multitask learning},
 pages = {41--75},
 volume = {28},
 number = {1},
 journal = {Machine Learning}
}

@misc{GregBrockman.2016,
 author = {{Greg Brockman} and {Vicki Cheung} and {Ludwig Pettersson} and {Jonas Schneider} and {John Schulman} and {Jie Tang} and {Wojciech Zaremba}},
 year = {2016},
 title = {OpenAI Gym}
}

\begin{IEEEbiography}[{\includegraphics[width=1in,height=1.25in,clip,keepaspectratio]{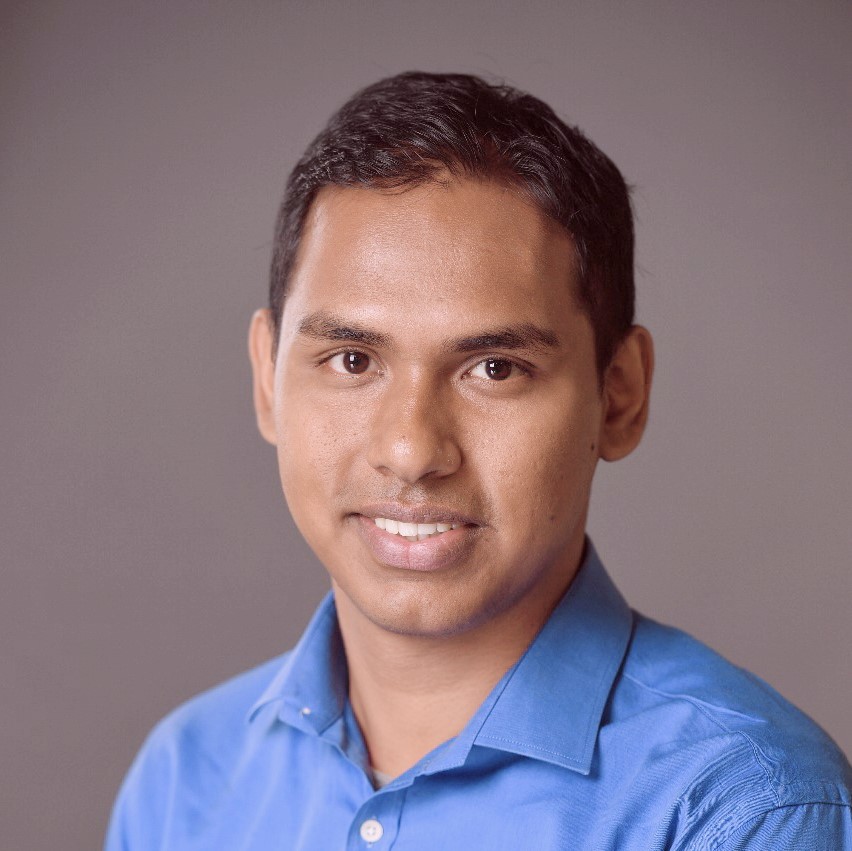}}]{Mohammed Sharafath Abdul Hameed}
received the M.Sc. degree in 2019 from the South Westphalia University of Applied Sciences, Soest, Germany. He is currently working as an research assistant at the department of automation technology at the South Westphalia University of Applied Sciences,Soest and working towards his PhD. His research interests include deep reinforcement learning, automation, production planning, and smart manufacturing.
\end{IEEEbiography}

\begin{IEEEbiography}[{\includegraphics[width=1in,height=1.25in,clip,keepaspectratio]{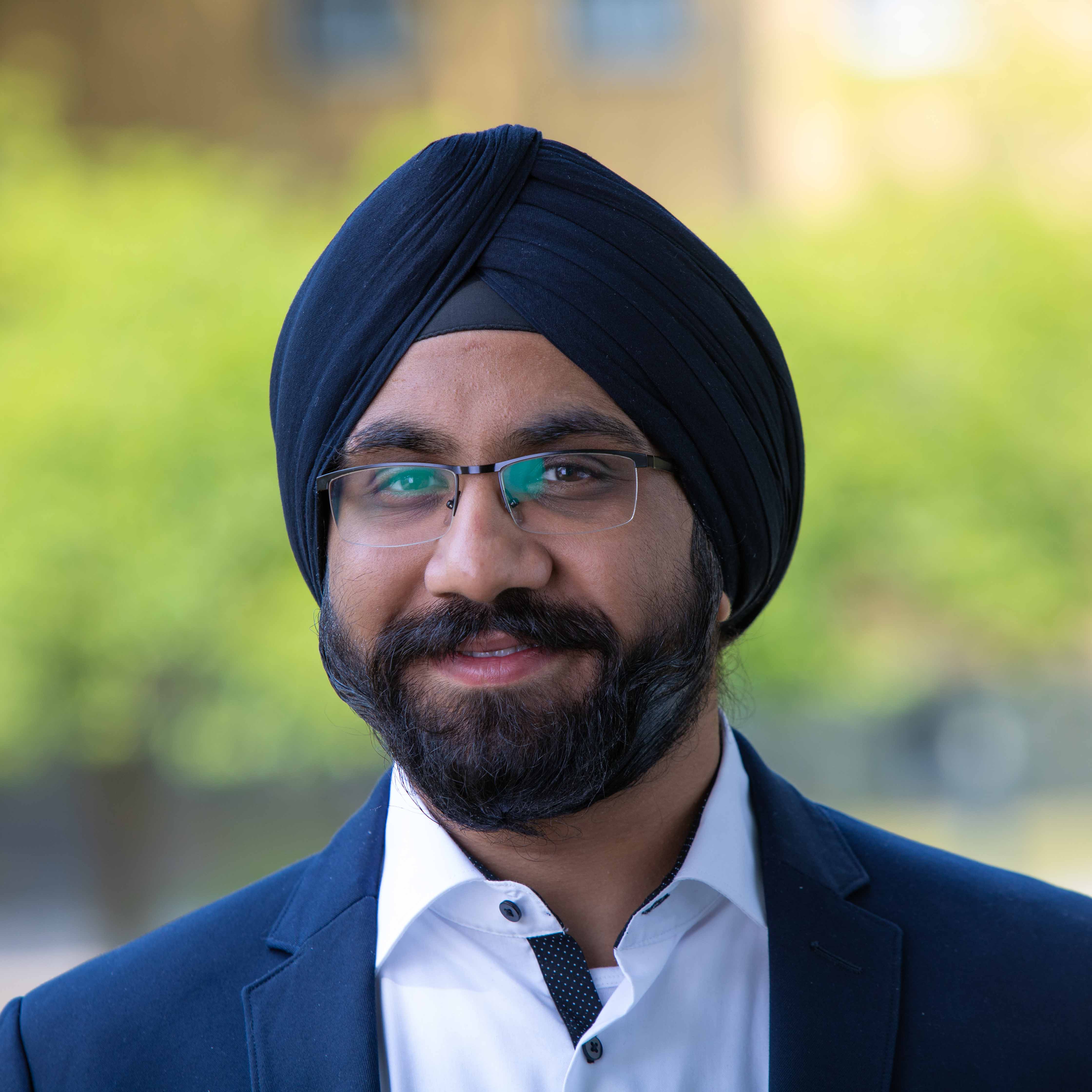}}]{Gavneet Singh Chadha}
received the M.Sc. degree in 2016 from the South Westphalia University of Applied Sciences, Soest, Germany. He is currently working as an research assistant at the department of automation technology at the South Westphalia University of Applied Sciences,Soest and working towards his PhD. His research interests include deep neural networks, fault diagnosis, predictive maintenance and machine learning.  
\end{IEEEbiography}

\begin{IEEEbiography}[{\includegraphics[width=1in,height=1.25in,clip,keepaspectratio]{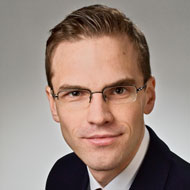}}]{Andreas Schwung}
received the Ph.D. degree	in electrical engineering from the Technische Universit\"at Darmstadt, Darmstadt, Germany, in 2011. From 2011 to 2015, he was an R\&D Engineer with MAN Diesel \& Turbo SE, Oberhausen, Germany. Since 2015, he has been a Professor of automation technology at the South Westphalia University of Applied Sciences, Soest, Germany. His research interests include model based control, networked automation systems, and intelligent data analytics with applications in manufacturing and process industry.
\end{IEEEbiography}

\begin{IEEEbiographynophoto}{Steven X. Ding}
received the Ph.D. degree in
electrical engineering from Gerhard Mercator
University of Duisburg, Duisburg, Germany,
in 1992.
From 1992 to 1994, he was an R\&D Engineer
with Rheinmetall GmbH. From 1995 to 2001, he
was a Professor of control engineering with the
University of Applied Science Lausitz, Senftenberg,
Germany, where he served as the Vice
President during 1998–2000. Since 2001, he
has been a Professor of control engineering and
the Head of the Institute for Automatic Control and Complex Systems
(AKS) with the University of Duisburg-Essen, Duisburg. His research
interests are model-based and data-driven fault diagnosis and fault tolerant
systems and their application in industry, with a focus on
automotive systems and mechatronic and chemical processes.

\end{IEEEbiographynophoto}

\end{document}